\documentclass[letterpaper]{article} % DO NOT CHANGE THIS
\usepackage{aaai24}  % DO NOT CHANGE THIS
\usepackage{times}  % DO NOT CHANGE THIS
\usepackage{helvet}  % DO NOT CHANGE THIS
\usepackage{courier}  % DO NOT CHANGE THIS
\usepackage[hyphens]{url}  % DO NOT CHANGE THIS
\usepackage{graphicx} % DO NOT CHANGE THIS
\usepackage{natbib}  % DO NOT CHANGE THIS AND DO NOT ADD ANY OPTIONS TO IT
\usepackage{caption} % DO NOT CHANGE THIS AND DO NOT ADD ANY OPTIONS TO IT
\usepackage{algorithm}
\usepackage{algorithmic}
\usepackage{amsmath}
\usepackage{amssymb}
\usepackage{multirow}
\usepackage{booktabs}
\usepackage{makecell}
\usepackage[table,xcdraw]{xcolor}
\usepackage{newfloat}
\usepackage{listings}
\DeclareCaptionStyle{ruled}{labelfont=normalfont,labelsep=colon,strut=off} % DO NOT CHANGE THIS
\floatstyle{ruled}
\newfloat{listing}{tb}{lst}{}
\floatname{listing}{Listing}
\nocopyright
\title{Context-I2W: Mapping Images to Context-dependent Words for Accurate Zero-Shot Composed Image Retrieval} 

\author {
    Yuanmin Tang\textsuperscript{\rm 1,\rm 2},
    Jing Yu\textsuperscript{\rm 1,\rm 2}\thanks{Corresponding author},
    Keke Gai\textsuperscript{\rm 3}, 
    Jiamin Zhuang\textsuperscript{\rm 1,\rm 2},
    Gang Xiong\textsuperscript{\rm 1},
    Yue Hu\textsuperscript{\rm 1}, 
    Qi Wu\textsuperscript{\rm 4}
}
\affiliations {
    \textsuperscript{\rm 1}Institute of Information Engineering, Chinese Academy of Sciences \\
    \textsuperscript{\rm 2}School of Cyber Security, University of Chinese Academy of Sciences \\
    \textsuperscript{\rm 3}Beijing Institute of Technology \\
    \textsuperscript{\rm 4}Australia Institute of Machine Learning, University of Adelaide \\
        \{tangyuanmin, yujing02, zhuangjiamin, xionggang, huyue\}@iie.ac.cn, gaikeke@bit.edu.cn, qi.wu01@adelaide.edu.au
}   

\usepackage{bibentry}
\begin{document}

\maketitle

\begin{abstract}
Different from Composed Image Retrieval task that requires expensive labels for training task-specific models, Zero-Shot Composed Image Retrieval (ZS-CIR) involves diverse tasks with a broad range of visual content manipulation intent that could be related to domain, scene, object, and attribute. The key challenge for ZS-CIR tasks is to learn a more accurate image representation that has adaptive attention to the reference image for various manipulation descriptions. In this paper, we propose a novel context-dependent mapping network, named Context-I2W,  for adaptively converting description-relevant Image information into a pseudo-word token composed of the description for accurate ZS-CIR. Specifically, an Intent View Selector first dynamically learns a rotation rule to map the identical image to a task-specific manipulation view. Then a Visual Target Extractor further captures local information covering the main targets in ZS-CIR tasks under the guidance of multiple learnable queries. The two complementary modules work together to map an image to a context-dependent pseudo-word token without extra supervision. Our model shows strong generalization ability on four ZS-CIR tasks, including domain conversion, object composition, object manipulation, and attribute manipulation. It obtains consistent and significant performance boosts ranging from 1.88\% to 3.60\% over the best methods and achieves new state-of-the-art results on ZS-CIR. Our code is available at \url{https://github.com/Pter61/context-i2w}.
\end{abstract}

\section{Introduction}

Given a reference image and a text description, Composed Image Retrieval (CIR) \cite{vo2019composing} aims to retrieve an image that is visually similar to a reference image while having visual modification according to the description. Unlike traditional content-based image retrieval \cite{datta2008image}, CIR is more flexible and accurate for users to express their search intent by incorporating both visual and language content, which brings emerging attention to internet search and e-commerce. Several CIR tasks have been proposed, including composing and manipulating objects for scene image search, attribute changes for fashion image search, and converting the image style for creative image search. There exist two core challenges of CIR: (1) accurately grounding the visual content in the reference image according to the description and (2) composing the relevant visual and textual content from different modalities for image retrieval. 

\begin{figure}[t]
    \centering
    \includegraphics[width=1.0\linewidth]{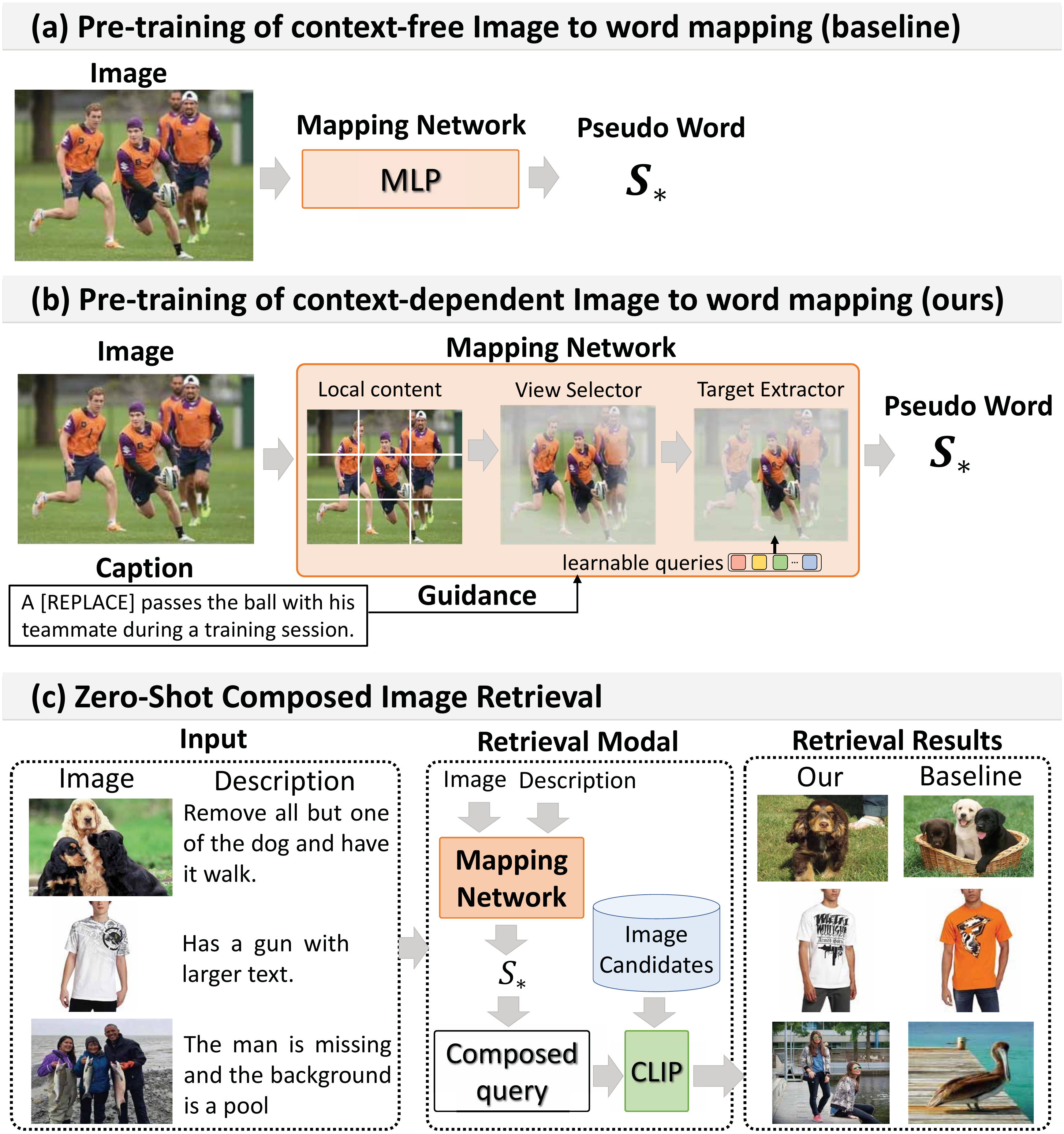}
    \caption{An illustration of our motivation. (a) Global visual mapping. (b) Our context-dependent visual mapping. (c) ZS-CIR process with the results of different mapping strategies.} 
    \label{fig:motivation}
\end{figure}

Several supervised approaches have been proposed to solve CIR problems \cite{Chen_2020_CVPR, Liu_2021_ICCV, Goenka_2022_CVPR, Baldrati_2022_CVPR}, which require a large amount of annotated triplets consisting of a reference image, a description, and a retrieved image. The supervised approaches trained on a specific task are also hard to generalize. To tackle these problems, the latest works \cite{Saito_2023_CVPR, baldrati2023zero} introduce the Zero-Shot Composed Image Retrieval (ZS-CIR) task. ZS-CIR aims to perform various CIR tasks without supervised training on task-specific triplet data. Current solutions treat the ZS-CIR task as a traditional text-based image retrieval problem, as illustrated in Figure \ref{fig:motivation}(c). They first learn a mapping network to convert the reference image embedding into a pseudo-word token, which is concatenated with the description to form a composed text query. Then, a pre-trained CLIP is leveraged to encode the query and the image candidates for text-to-image retrieval in a zero-shot mode. However, these models convert the whole visual information of an image into the same pseudo-word for different descriptions, as shown in Figure \ref{fig:motivation}(a). This limits the model's flexibility of adaptively selecting and mapping the visual information. In fact, only part of the visual content is relevant to the manipulation intent. Considering the queries in Figure \ref{fig:motivation}(c), existing approaches are vulnerable when the description manipulates a certain object from multiple ones (\textit{e.g.,} one of all the dogs), partial attributes of the whole object (\textit{e.g.,} text size on the T-shirt) or the background/foreground of the global image (\textit{e.g.,} the sea in the background). 

In this work, we regard a description as the context of a reference image and propose a \textit{\textbf{Context}-dependent mapping network to adaptively convert description-relevant \textbf{I}mage information into a pseudo \textbf{W}ord} (\textbf{Context-I2W}) for accurate ZS-CIR. To be generic for diverse CIR tasks, Context-I2W adaptively selects description-relevant information from the image in a hierarchical mode as illustrated in Figure \ref{fig:motivation}(b): the Intent View Selector module first learns various mapping rules to dynamically map the visual embeddings to different views in a context-dependent way. Thus, each view captures the visual information from the task-specific manipulation intent (\textit{e.g.,} domain, scene, and attribute) for even identical input. Then, the Visual Target Extractor module further collects local information from distinct aspects (\textit{e.g.,} foreground, background, objects, and details) under the guidance of multiple learnable queries. The two modules work together to map an image to a context-dependent pseudo-word token. Independent of expensive description-region labels, Context-I2W is trained with a contrastive loss between the retrieved image and the pseudo-word enhanced composed query. 

The main contributions are summarized as follows: (1) We propose a novel image to context-dependent word mapping network augmented by view selection and target extraction. We consider manipulation descriptions and learnable queries as multi-level constraints for visual information filtering, which sheds new light on the vision-to-language alignment mechanism. (2) The proposed mapping network of Context-I2W is incorporated into the ZS-CIR framework and demonstrated beneficial to tackle the existing challenges of foreground/background differentiation, multiple object combination, and fine-grained image editing. Context-I2W outperforms context-free mapping approaches and most supervised solutions. (3) Our Context-I2W mapping approach is consistently effective for diverse ZS-CIR tasks. It significantly improves CIR from 1.88\% to 3.60\% across four CIR tasks, including domain conversion, object composition, object/scene manipulation, and attribute manipulation. It establishes new state-of-the-art results and further impacts a broader range of vision and language applications.

\section{Related Work}
\subsubsection{Composed Image Retrieval.} Composed Image Retrieval (CIR) integrates image and text for retrieval \cite{Vo_2019_CVPR}. Current models use late-fusion techniques based on separately extracted visual and language features to merge them for retrieval \cite{Baldrati_2022_CVPR, Liu_2021_ICCV, Chen_2020_CVPR, Shi_2023_ICCV}. Another kind of approach \cite{Goenka_2022_CVPR, 10.1007/978-3-031-19833-5_37} incorporates image, text, and tag features through pre-trained models.  CLIP4Cir \cite{Baldrati_2022_CVPR} achieves state-of-the-art by fine-tuning a CLIP \cite{radford2021learning} text encoder and employing the Combiner late-fusion module but still relies on annotated triplets. In contrast, Pic2Word \cite{Saito_2023_CVPR} and SEARLE \cite{baldrati2023zero} are zero-shot CIR models trained on common image-text pairs without costly annotated CIR datasets. Pic2Word aligns the whole image features with text features, while SEARLE further integrates a pseudo-word token with a GPT-based \cite{brown2020language} caption for data generation. However, these methods ignore the context clues for mapping description-desired visual content, thus introducing much irrelevant noise for query composition and resulting in incorrect results. To address this, we propose a context-dependent word mapping strategy, enabling the text encoder to access essential image features for accurate retrieval.

\begin{figure*}
    \centering
    \includegraphics[width=1.0\linewidth]{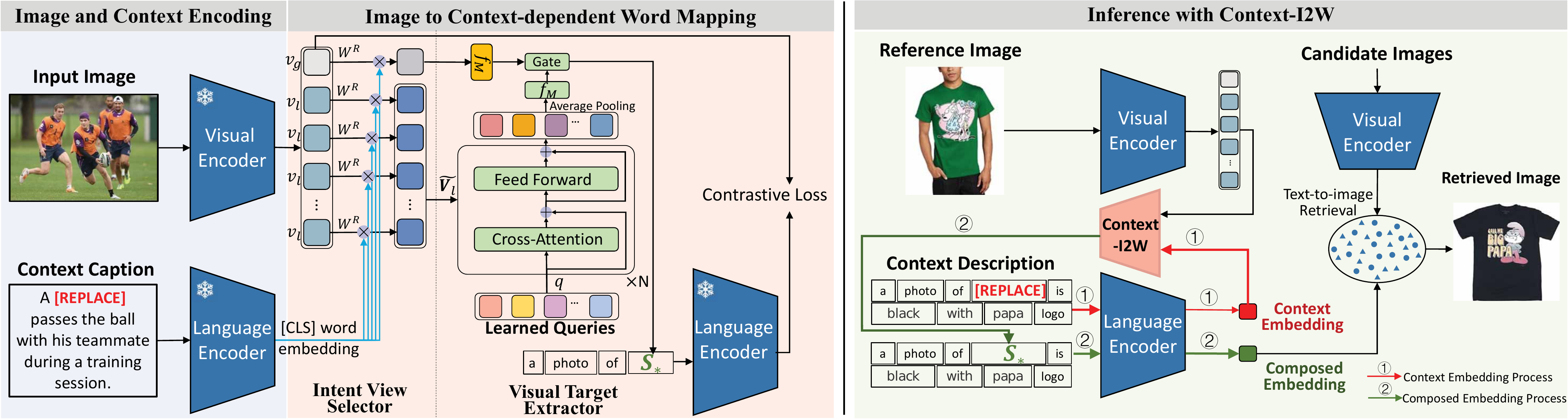}   
    \caption{An overview of our Context-I2W model. \textbf{Pre-training (left):} Image to Context-dependent Word Mapping aims to extract caption-relevant visual content from the view level to the target level and map it to a pseudo-word token $S_*$. \textbf{Inference (right):} Map the inference image to $S_*$ and form the composed query in a unified language space for ZS-CIR.}
    \label{fig:model-architecture}
\end{figure*}

\subsubsection{Vision and Language Pre-training Models.} Typical Vision and Language Pre-training (VLP) models like CLIP \cite{radford2021learning} and and ALIGN \cite{NEURIPS2021_50525975} are trained on large-scale image-text pairs, which obtain pre-trained knowledge of implicit alignment between images and texts. Recent advance in VLP \cite{ Zhou_2022_CVPR, song2022clip} explores frozen VLP models to map the encoded image features and text features to the aligned semantic space for various zero-shot tasks \cite{pmlr-v162-li22n,song2022clip}. In this work, we utilize the pre-trained CLIP as the mapping network backbone with improvement by adaptively selecting description-relevant visual information for image-text alignment. Besides, recent works \cite{NEURIPS2022_960a172b, li2023blip2} inspired by DETR \cite{carion2020end} utilize learnable queries for image information selection, leading to performance boost by capturing targeted information before prediction. Our work defines multiple learnable queries as guidance to capture specific parts of visual content, which provides explainable clues and fine-grained visual features for more accurate ZS-CIR.

\subsubsection{Mapping Image as One Word.} Several methods \cite{li2020oscar, zhang2021vinvl} aim to represent image regions as word tokens via VLP models, which rely on the effectiveness of object detectors. However, ZR-CIR tasks extend the alignment ability beyond objects to scenes, styles, attributes, \textit{ect}. Our Context-I2W addresses this issue by using patch-based image features instead of object features to align description-relevant visual content to the word token space dynamically. PALAVRA \cite{10.1007/978-3-031-20044-1_32} addresses personalized image retrieval through cycle contrastive loss, which needs class-wise and caption annotations. Our proposed model achieves fine-grained image-to-word mapping without the need of any extra annotations. Some other methods \cite{Kumari_2023_CVPR, mokady2021clipcap, zhu2023visualize, tam2023simple} use one word token to represent multiple images of the same object for text-to-image generation.  Our model doesn't require expensive image grouping or supervised training.

\section{Methodology}

Given a reference image $I$ and a manipulation description $T$, ZS-CIR aims to retrieve images from an image database that are visually similar to $I$ while having visual modification required in $T$. Figure \ref{fig:model-architecture} gives a detailed illustration of our model. We first introduce a new approach of encoding context based on pre-trained CLIP \cite{radford2021learning}, which contains vision-language alignment knowledge to provide contextual constraints for visual information selection. This process is consistent for both contextual caption in the pre-training stage and contextual description in the reference stage. Then we learn a mapping network of Context-I2W to convert the reference image $I$ into a pseudo-word token $S_{*}$, which is not an actual word but a representation of the image in the word token embedding space. In this work, $S_{*}$ accurately depicts the visual content of context-guided manipulation intention (Intent View Selector module) and target-attended manipulation details (Visual Target Extractor module) specified in $T$. To effectively compose  $I$ and $T$ across different modalities for zero-shot image retrieval, we construct a composed query in the form of a sentence $P$ ``a photo of $S_{*}$ that {$T$}” and embed it using the frozen text encoder of CLIP. Given the composed query embedding, we embed each candidate image $I_{c}$ by the frozen image encoder of CLIP and regard ZS-CIR as a traditional text-to-image retrieval task by measuring the similarity between $P$ and $I_{c}$.

\subsection{Image and Context Encoding}
Since the pre-trained vision-language models are strong at modeling the cross-modal implicit alignment, we first utilize the pre-trained CLIP model to encode the context and image for fine-grained image-to-word mapping. We apply the visual encoder of frozen CLIP to represent the reference image $I$ by a set of visual feature vectors  $\boldsymbol{V}$ =$\{\boldsymbol{v}_i\}_{i=1}^m\subseteq{\mathbb{R}^{d\times{m}}}$ ($m=257$, $d=1024$), where $\boldsymbol{v}_1$ denotes the global image feature $\boldsymbol{v}_g$ while other ones denote local patch features $\boldsymbol{V}_l$ =$\{\boldsymbol{v}_{i}\}_{i=2}^m$. In this work, we have a dataset of image-caption pairs for learning the Context-I2W network and a set of image-description pairs for ZS-CIR. We denote both captions and descriptions identically as the context of the corresponding images. Since the context generally introduces a visual target (\textit{e.g.,} rugby player or T-shirt in Figure \ref{fig:model-architecture}) by describing its relevant information  (\textit{e.g.,} surroundings or attributes). We represent the context by target-relevant information except for the target for two goals. First, we need to ensure that the context captures the described intent view of the visual content. In other words, the context features serve as mapping rules that convert the identical image representation to a text-specific view, such as the visual aspects of the surrounding scene, image domain, visual attributes, etc. Second, we need to ensure that such context features can integrate with the visual target feature in a complementary way for accurate pseudo-token mapping. To this end, we extract the first subject term in the context by a part-of-speech tagger \textit{spacy} \cite{honnibal2020spacy} and replace it with a learnable token \texttt{[REPLACE]}. We then feed the rewritten sentence to the visual encoder of frozen CLIP and obtain the \texttt{[CLS]} token embedding $\boldsymbol{t}=\{{t_i}\}_{i=1}^d\in\mathbb{R}^{d\times{1}}$ as visual feature extraction guidance.   

\subsection{Image to Context-dependent Word Mapping}

Since each context within an image-context pair contains a view-different and detail-specific description, we propose two modules to progressively learn the context-relevant visual features to map it to a context-dependent pseudo-token for accurate ZS-CIR. The two modules constrain the visual representation from complementary aspects: the \textit{Intent View Selector} (IVS for short) preserves the context-desired view of visual features from the identical image features by context-guidance feature rotation. The \textit{Visual Target Extractor}  (VTE for short) further aggregates visual features of context-specific targets from the desired view features. Both the target and its relevant information are adaptively combined and mapped into the word token space by cross-modal contrastive learning.   

\subsubsection{Intent View Selector.} Based on the encoded image and context embeddings, this module aims to represent the visual content from the view of intent described in the context without interfering with the visual content. From a mathematical point of view, the context serves by rotating the image embedding space according to the context constraint, thereby selecting the view representation of $\boldsymbol{V}$ without changing its content. Specifically, we feed each visual embedding $\boldsymbol{v}_i$ to a fully-connected layer with learnable weight matrix ${\boldsymbol{W}}^{R}= \{{\boldsymbol{w}}^{R}_k\}_{k=1}^d\subseteq{\mathbb{R}^{d\times{d}}}$. Each column of ${\boldsymbol{W}}^{R}$ is normalization with $\|{\boldsymbol{w}}^{R}_k\|=1$. The operation of the IVS is formulated as: 
\begin{equation}
\begin{aligned}
\tilde{\boldsymbol{v}}_i &= ({(\boldsymbol{W}^{R})}^{\textsuperscript{T}}\boldsymbol{v}_i)\odot{\boldsymbol{t}} \\
                  &= [t_1\|\boldsymbol{v}_i\|\|\boldsymbol{w}^{R}_1\|\cos{\theta_1}, ..., t_d\|\boldsymbol{v}_i\|\|\boldsymbol{w}^{R}_d\|\cos{\theta_d}]^{\textsuperscript{T}} \\
                  &= [t_1\|\boldsymbol{w}^{R}_1\|\cos{\theta_1}, ..., t_d\|\boldsymbol{w}^{R}_d\|\cos{\theta_d}]^{\textsuperscript{T}}\|\boldsymbol{v}_i\| \\
                  &= [t_1\cos{\theta_1}, ..., t_d\cos{\theta_d}]^{\textsuperscript{T}}\|\boldsymbol{v}_i\| 
\end{aligned}
\label{f:Walign}
\end{equation}
\noindent where $\odot$ represents element-wise multiplication and $\theta_k$ is the angle between $\boldsymbol{w}^{R}_k$ and $\tilde{\boldsymbol{v}}_i$. Given the visual feature $\boldsymbol{v}_i$ and the fixed $\boldsymbol{W}^{R}$, the selected visual feature $\tilde{\boldsymbol{v}}_i$ is only affected by the context $\boldsymbol{t}$. Furthermore, we normalize $\boldsymbol{t}$ by $\sqrt{\sum_{k=1}^{d}\left(t_{k}\cos \theta_{k}\right)^{2}}$ in Equation \ref{f:Walign} and thus ``rotate'' $\boldsymbol{v}_i$ in the feature space without changing the information content. That is, we represent the visual feature from the intent view required in context while keeping the information content unchanged, which is essential for providing comprehensive information to extract the visual target further.  

\subsubsection{Visual Target Extractor.}  Given the visual features from a specific intent view, the AI agent needs to focus further on diverse partial content in the image with respect to the context. The contextual descriptions mostly attend to several typical visual targets, including global information in domain conversion, foreground or background in scene manipulation, specific objects in object composition, visual details in attribute manipulation, \textit{etc}. To adaptively capture the desired visual content, we define a set of learnable query embeddings  $\boldsymbol{X}=\{\boldsymbol{x}_k\}_{k=1}^n\in{\mathbb{R}^{d\times{n}}}$, where $d$ is the embedding dimension and $n$ is the number of learnable queries. Each learnable query $\boldsymbol{x}_k$ represents a kind of context-mentioned visual target. As illustrated in Figure \ref{fig:model-architecture}(left), we apply cross-attention to gather target-relevant visual information from all the local patch features $\tilde{\boldsymbol{V}}_l = \{\tilde{\boldsymbol{v}}_i\}_{i=2}^m$ for the learnable queries $\boldsymbol{X}$ as follows. First, we compute the query, key and value through linear projections, \textit{i.e.,} $\boldsymbol{Q} = \boldsymbol{X}\boldsymbol{W}^Q$, $\boldsymbol{K} = [\boldsymbol{X}, \tilde{\boldsymbol{V}}_l]\boldsymbol{W}^K$, $\boldsymbol{V} = [\boldsymbol{X}, \tilde{\boldsymbol{V}}_l]\boldsymbol{W}^V$.  $[\boldsymbol{X}, \tilde{\boldsymbol{V}}_l]$ denotes concatenating the two matrices, which enhances the interaction between learnable queries and local patches with better performance. Then, the learnable queries from the current cross-attention block $\boldsymbol{X}^{i}$ is calculated as:
\begin{gather}
\boldsymbol{X}_{att}^{i} = \operatorname{Att}(\boldsymbol{Q}, \boldsymbol{K}, \boldsymbol{V})=\textit{softmax}\left(\frac{\boldsymbol{Q K}^{\top}}{\sqrt{d}}\right)\boldsymbol{V} \\
\boldsymbol{X}^{i} = \operatorname{FFW}(\boldsymbol{X}_{att}^{i} + \boldsymbol{X}^{i-1}) + \boldsymbol{X}_{att}^{i}
\label{f:attn}
\end{gather}
\noindent where $\boldsymbol{X}^{i-1}$ are learnable queries from the previous block and $\operatorname{FFW}(\cdot)$ denotes 2-layer feed-forward networks. When mapping the visual content to a pseudo-word token, both the target and relevant visual content are complementary to form the complete context information. We design a learnable scalar $gate$ to decide the contribution of the target content and integrate the two parts to form the final pseudo-word token embedding $\boldsymbol{S}_{*}$ as follows:  
\begin{gather}
\boldsymbol{S}_{*} = f_{M_l}(gate\cdot\operatorname{Avg}(\boldsymbol{X}_{output}))+ f_{M_g}(\tilde{\boldsymbol{v}}_{g})
\label{f:gate}
\end{gather}

\noindent where $\boldsymbol{X}_{output}$ is the output query embeddings from $N$ transformer blocks, $\operatorname{Avg}(\cdot)$ denotes average pooing $f_{M_l}(\cdot)$ and $f_{M_g}(\cdot)$ respectively denote local and global mapping of 3-layer feed-forward networks. 

\subsubsection{Cross-modal Contrastive Loss.} 

To map the pseudo token $\boldsymbol{S}_{*}$ to the word token space,  we first replace the special token \texttt{[REPLACE]} in a prompt sentence ``\texttt{a photo of [REPLACE]}'' with $\boldsymbol{S}_{*}$ and feed it to the language encoder of CLIP to obtain the sentence embedding $\boldsymbol{t}_s$. We aim to match an image to its paired context-dependent prompt sentence while separating unpaired
ones. We minimize the symmetric contrastive loss between the global visual embedding $\boldsymbol{v}_g$ and the prompt sentence embedding $\boldsymbol{t}_s$ as follows:
\begin{equation}
\begin{aligned}
\mathcal{L}=\mathcal{L}_{t 2 i}(\boldsymbol{t}_s,\boldsymbol{v}_g)+\mathcal{L}_{i 2 t}(\boldsymbol{t}_s, \boldsymbol{v}_g)
\end{aligned}
\label{f:loss}
\end{equation}

\begin{table*}[t]
\centering
\scalebox{1.05}
{\scriptsize
\begin{tabular}{cccccccccccc}
\toprule
                           &              & \multicolumn{2}{c}{Cartoon}                       & \multicolumn{2}{c}{Origami}                        & \multicolumn{2}{c}{Toy}                            & \multicolumn{2}{c}{Sculpture}                      & \multicolumn{2}{c}{Average}   \\  \cmidrule(lr){3-4}\cmidrule(lr){5-6}\cmidrule(lr){7-8}\cmidrule(lr){9-10}\cmidrule(lr){11-12}
                           %\cline{3-10}
Supervision                & Methods      & R10          & R50                                & R10           & R50                                & R10           & R50                                & R10           & R50                                & R10           & R50           \\ \hline
\multirow{5}{*}{ZERO-SHOT} & Image-only   & 0.3          & \multicolumn{1}{c|}{4.5}           & 0.2           & \multicolumn{1}{c|}{1.8}           & 0.6           & \multicolumn{1}{c|}{5.7}           & 0.3           & \multicolumn{1}{c|}{4.0}           & 0.4           & 4.0           \\
                           & Text-only    & 0.2          & \multicolumn{1}{c|}{1.1}           & 0.8           & \multicolumn{1}{c|}{3.7}           & 0.8           & \multicolumn{1}{c|}{2.4}           & 0.4           & \multicolumn{1}{c|}{2.0}           & 0.5           & 2.3           \\
                           & Image+Text   & 2.2          & \multicolumn{1}{c|}{13.3}          & 2.0           & \multicolumn{1}{c|}{10.3}          & 1.2           & \multicolumn{1}{c|}{9.7}           & 1.6           & \multicolumn{1}{c|}{11.6}          & 1.7           & 11.2          \\
                           & Pic2Word (CVPR 2023)    & 8.0          & \multicolumn{1}{c|}{21.9}          & 13.5          & \multicolumn{1}{c|}{25.6}          & 8.7           & \multicolumn{1}{c|}{21.6}          & 10.0          & \multicolumn{1}{c|}{23.8}          & 10.1          & 23.2          \\ 
                           & \textbf{Context-I2W} & \textbf{10.2} & \multicolumn{1}{c|}{\textbf{26.1}} & \textbf{17.5} & \multicolumn{1}{c|}{\textbf{28.7}} & \textbf{11.6} & \multicolumn{1}{c|}{\textbf{27.4}} & \textbf{12.1} & \multicolumn{1}{c|}{\textbf{28.2}} & \textbf{12.9} & \textbf{27.6} \\ \cmidrule(lr){1-12}
CIRR                       & Combiner (CVPR 2022)    & 6.1          & \multicolumn{1}{c|}{14.8}          & 10.5          & \multicolumn{1}{c|}{21.3}          & 7.0           & \multicolumn{1}{c|}{17.7}          & 8.5           & \multicolumn{1}{c|}{20.4}          & 8.0           & 18.5          \\
Fashion-IQ                 & Combiner (CVPR 2022)    & 6.0          & \multicolumn{1}{c|}{16.9}          & 7.6           & \multicolumn{1}{c|}{20.2}          & 2.7           & \multicolumn{1}{c|}{10.9}          & 8.0           & \multicolumn{1}{c|}{21.6}          & 6.0           & 17.4          \\ \bottomrule
\end{tabular}}
\caption{Results on ImageNet for domain conversion.}
\label{tab:imgnet}
\end{table*}

The two contrastive loss terms with a temperature hyper-parameter $\tau$ that controls the strength of penalties on hard negative samples are defined as:
\begin{equation}
\begin{aligned}
\mathcal{L}_{t 2 i}(\hat{\boldsymbol{t}}_s, \hat{\boldsymbol{v}}_g)=-\frac{1}{|\mathcal{B}|} \sum_{i \in \mathcal{B}} \log \frac{\exp \left(\tau \hat{\boldsymbol{t}}_{s_{i}}^{T} \hat{\boldsymbol{v}}_{g_{i}}\right)}{\sum_{j \in \mathcal{B}} \exp \left(\tau \hat{\boldsymbol{t}}_{s_{i}}^{T} \hat{\boldsymbol{v}}_{g_{j}}\right)}
\end{aligned}
\label{f:CTL_1}
\end{equation}
\begin{equation}
\begin{aligned}
\mathcal{L}_{i 2 t}(\hat{\boldsymbol{t}}_s, \hat{\boldsymbol{v}}_g)=-\frac{1}{|\mathcal{B}|} \sum_{i \in \mathcal{B}} \log \frac{\exp \left(\tau \hat{\boldsymbol{v}}_{g_{i}}^{T} \hat{\boldsymbol{t}}_{s_{i}}\right)}{\sum_{j \in \mathcal{B}} \exp \left(\tau \hat{\boldsymbol{v}}_{g_{i}}^{T} \hat{\boldsymbol{t}}_{s_{j}}\right)}
\end{aligned}
\label{f:CTL_2}
\end{equation}
\noindent where $\hat{\boldsymbol{t}}_{s_{i}} = \frac{\boldsymbol{t}_{s_{i}}}{\|\boldsymbol{t}_{s_{i}}\|}$ and $\hat{\boldsymbol{v}}_{g_{j}} = \frac{\boldsymbol{v}_{g_{j}}}{\|\boldsymbol{v}_{g_{j}}\|}$ are the normalized features of $i$-th prompt sentence embedding $\boldsymbol{t}_{s_i}$ and the $j$-th image global embedding $\boldsymbol{v}_{g_{j}}$ in a batch $\mathcal{B}$.

\subsection{Inference with Context-I2W}
\label{sec:inference}
In the inference stage, we compose the reference image with the paired context description and compare the composed query with candidate images for retrieval. As shown in Figure \ref{fig:model-architecture} (right), we first modify the context description with the learnt  \texttt{[REPLACE]} token to form a prompt sentence and feed it to the pre-trained language encoder of CLIP and subsequently the Context-I2W network. Then we obtain the mapped pseudo-word token embedding $S_*$ with the context-dependent visual information. Since the pseudo-word token is in the same space as actual words, we replace the \texttt{[REPLACE]} token with $S_*$ in the prompt sentence to form a composed query. The composed query as well as each candidate image, is encoded by CLIP for similarity measurement. We rank the candidates based on their similarity scores. 

Since our focus is on studying the context-dependent word mapping for ZS-CIR, we utilize the same prompt in the most recent work \cite{Saito_2023_CVPR} for a fair comparison. We show prompt examples for different ZS-CIR tasks: \textbf{(a) Domain conversion} aims to modify the domain of the reference image. The prompt is defined as \texttt{a [domain tag] of [REPLACE]}; \textbf{(b) Object composition} retrieves an image that contains an object in the reference image and other object tags. The prompt is in the format of \texttt{a photo of [REPLACE], [obj$_1$ tag] and [obj$_2$ tag], $\dots$, and [obj$_n$ tag]}; \textbf{(c) Sentence manipulation} modifies the reference image based on a sentence. We simply append the sentence with the special token as  \texttt{a photo of [REPLACE], [sentence]}.

\section{Experiments}

\begin{figure}[t]
    \centering
    \centering
    \includegraphics[width=1.0\linewidth]{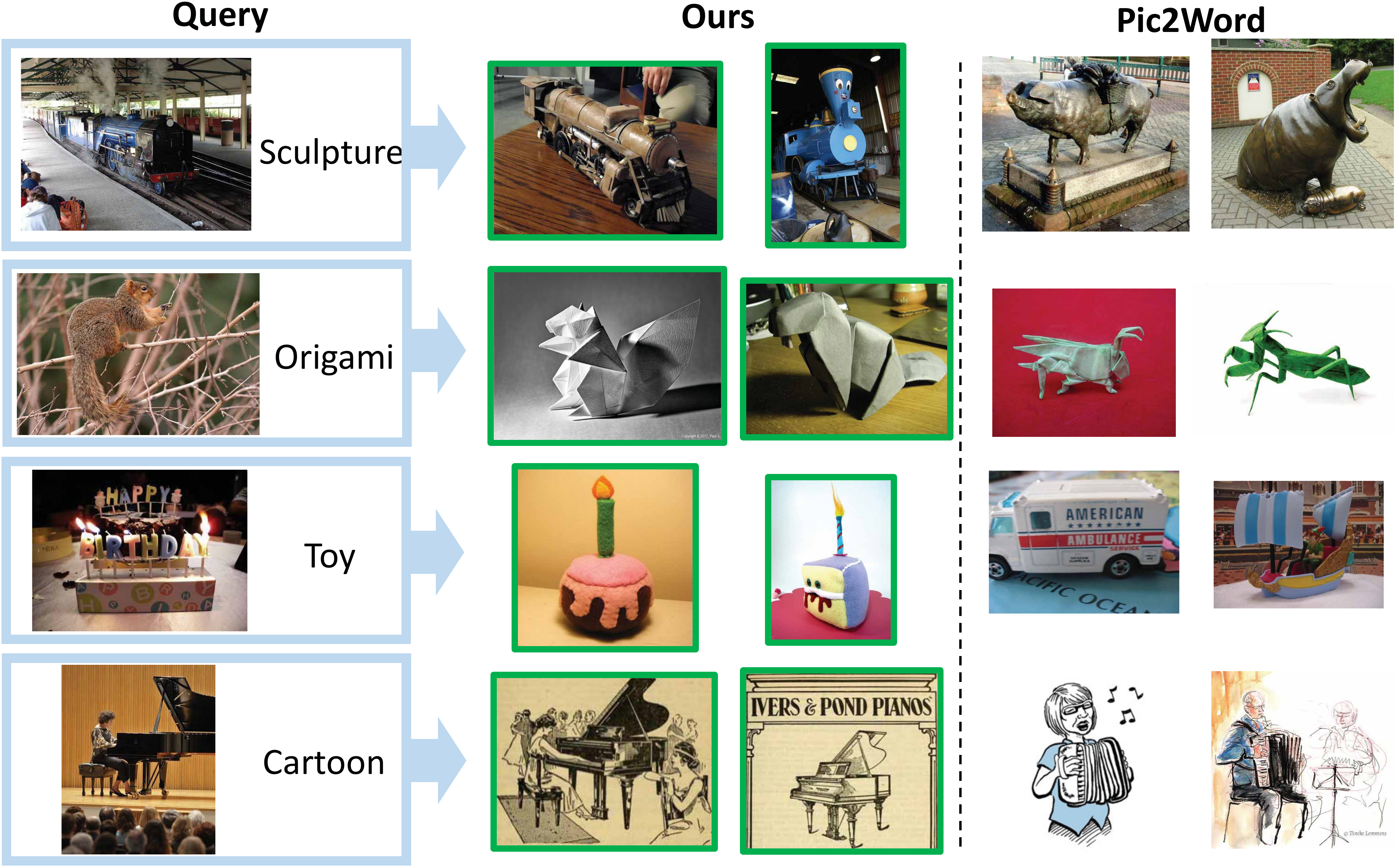}
    \caption{Retrieved results on the domain conversion task.}
    \label{fig:imgnet}
\end{figure}

\noindent \textbf{Datasets.} We evaluate our model on four ZS-CIR datasets, \textit{i.e.,} COCO \cite{10.1007/978-3-319-10602-1_48} for object composition, ImageNet \cite{deng2009imagenet, Hendrycks_2021_ICCV} for domain conversion, CIRR \cite{Liu_2021_ICCV} for object/scene manipulation, and Fashion-IQ \cite{Wu_2021_CVPR} for attribute manipulation. All the dataset settings follow the recent works \cite{Saito_2023_CVPR, baldrati2023zero} for a fair comparison. 

\noindent (1) Domain conversion.  This dataset comprises 16,983 images of 200 classes from four domains, \textit{i.e.,} cartoon, origami, toy, and sculpture. We use the prompt (a) in inference. \noindent (2) Object composition. The dataset contains images with corresponding lists of object labels and instance masks of query images. We randomly crop one object and mask its background using its instance mask to create a reference image. We use the prompt (b) in inference. \noindent (3) Object/scene manipulation. A reference image is an instruction for manipulating an object or the background scene. We apply the prompt (c) in inference. \noindent (4) Attribute manipulation.  This dataset includes various description sentences for manipulating image attributes. We utilize the prompt (c) in inference. 

\noindent \textbf{Implementation Details.} We adopt ViT-L/14 CLIP \cite{radford2021learning} pre-trained on 400M image-text paired data. For training Context-I2W, we utilize the Conceptual Caption dataset \cite{DBLP:conf/acl/SoricutDSG18}, which comprises 3M images. The number of cross-attention blocks is $6$. The number of learnable queries is $4$. To improve training stability, we initialize the learnable scalar of tanh-gating to 0 \cite{bachlechner2021rezero}. We employ AdamW \cite{loshchilov2018decoupled} with a learning rate of $5\times10^{-5}$, weight decay of $0.1$, and a linear warmup of $10000$ steps. The batch size for contrastive learning is $1024$. Our model is trained on $4$ Tesla V100 (32G) GPUs for $24$ hours. To ensure reliable results, we report the performance averaged over three trials.

\subsection{Quantitative and Qualitative Results}
We compare Context-I2W with several ZS-CIR methods, including: 1) \textbf{Text-only}: the similarity is computed based on the CLIP features of the descriptions and the candidate images; 2) \textbf{Image-only}: retrieves the most similar images to the reference image via CLIP visual features; 3) \textbf{Image + Text}: the summation of CLIP features of the reference image and the description; 4)  \textbf{Pic2Word} \cite{Saito_2023_CVPR}: maps the entire visual features of the reference image into a pseudo-word token within the CLIP token embedding space; 5) \textbf{SEARLE-XL} \cite{baldrati2023zero}: Similar to Pic2Word, it further integrates the pseudo-word token with the relative caption generated by GPT \cite{brown2020language} and distill for efficiency.
We show the reported results of SEARLE-XL on CIRR and Fashion-IQ. Also, we compare Context-I2W with the published results of widely compared supervised models, including Combiner \cite{Baldrati_2022_CVPR} (Combiner$^*$  in tables indicates using ResNet50x4 as a backbone), TIRG \cite{vo2019composing}, ARTEMIS \cite{delmas2022artemis}, CIRPLANT\cite{Liu_2021_ICCV} and MAAF \cite{dodds2020modality}.

\begin{table}[t]
\centering
\scalebox{1.05}
{\scriptsize
\begin{tabular}{cccccccccccc}
\toprule
Supervision                                     & Methods                           & R1            & R5            & \multicolumn{1}{l}{R10} \\  \midrule
\multicolumn{1}{c|}{\multirow{5}{*}{ZERO-SHOT}} & \multicolumn{1}{c|}{Image-only}   & 8.6           & 15.4          & 18.9                    \\
\multicolumn{1}{c|}{}                           & \multicolumn{1}{c|}{Text-only}    & 6.1           & 15.7          & 23.5                    \\
\multicolumn{1}{c|}{}                           & \multicolumn{1}{c|}{Image+Text}   & 10.2          & 20.2          & 26.6                    \\
\multicolumn{1}{c|}{}                           & \multicolumn{1}{c|}{Pic2Word}     & 11.5          & 24.8          & 33.4                    \\
\multicolumn{1}{c|}{}                           & \multicolumn{1}{c|}{\textbf{Context-I2W}} & \textbf{13.5}          & \textbf{28.5}          & \textbf{38.1}           \\ \cmidrule(lr){1-5}
\multicolumn{1}{c|}{CIRR}                       & \multicolumn{1}{c|}{Combiner}     & 9.9           & 22.8          & 32.2                    \\
\multicolumn{1}{c|}{Fashion-IQ}                 & \multicolumn{1}{c|}{Combiner}     & 13.2          & 27.1          & 35.2                   \\ \bottomrule
\end{tabular}}
\caption{Results of the object composition task using COCO.}
\label{tab:coco}
\end{table}

\begin{table}[t]
\centering
\scalebox{1.05}
{\scriptsize
\begin{tabular}{cccccccccccc}
\toprule
Supervision                                      & Methods                                                   & R1                                               & R5                                               & \multicolumn{1}{l}{R10}                          & R50                                              \\ \midrule
\multicolumn{1}{c|}{}                            & \multicolumn{1}{c|}{Image-only}                           & 7.4                                              & 23.6                                             & 34.0                                             & 57.4                                             \\
\multicolumn{1}{c|}{}                            & \multicolumn{1}{c|}{Text-only}                            & 20.9                                             & 44.8                                             & 56.7                                             & 79.1                                             \\
\multicolumn{1}{c|}{}                            & \multicolumn{1}{c|}{Image+Text}                           & 12.4                                             & 36.2                                             & 49.1                                             & 78.2                                             \\
\multicolumn{1}{c|}{}                            & \multicolumn{1}{c|}{Pic2Word}                             & 23.9                                            & 51.7                                           & 65.3                                            & 87.8                                            \\
\multicolumn{1}{c|}{}                            & \multicolumn{1}{c|}{SEARLE-XL}                   & 24.2                                            & 52.4                                            & 66.3                                            & 88.6                                   \\
\multicolumn{1}{c|}{\multirow{-6}{*}{ZERO-SHOT}}                             & \multicolumn{1}{c|}{\textbf{Context-I2W}}                         & \textbf{25.6}                                   & \textbf{55.1}                                   & \textbf{68.5}                                   & \textbf{89.8}                                   \\
\cmidrule(lr){1-6}
\multicolumn{1}{c|}{CIRR}                        & \multicolumn{1}{c|}{Combiner}                             & 30.3                                             & 60.4                                             & 73.2                                             & 92.6                                             \\
\multicolumn{1}{c|}{Fashion-IQ}                  & \multicolumn{1}{c|}{Combiner}                             & 20.1                                             & 47.7                                             & 61.6                                             & 85.9                                             \\ %\cmidrule(lr){1-6}
\multicolumn{1}{c|}{CIRR}                        & \multicolumn{1}{c|}{Combiner*}                            & 33.6                                             & 65.4                                             & 77.4                                             & 95.2                                             \\
\multicolumn{1}{c|}{CIRR}                        & \multicolumn{1}{c|}{TIRG}                                 & 14.6                                             & 48.4                                             & 64.1                                             & 90.0                                             \\
\multicolumn{1}{c|}{CIRR}                        & \multicolumn{1}{c|}{ARTEMIS}                              & 17.0                                             & 46.1                                             & 61.3                                             & 87.7                                             \\
\multicolumn{1}{c|}{CIRR}                        & \multicolumn{1}{c|}{CIRPLANT}                             & 19.6                                             & 52.6                                             & 68.4                                             & 92.4                                             \\ \bottomrule
\end{tabular}}
\caption{Results on CIRR for object composition.}
\label{tab:CIRR}
\end{table}

\begin{figure}[t]
    \centering
    \centering
    \includegraphics[width=1.0\linewidth]{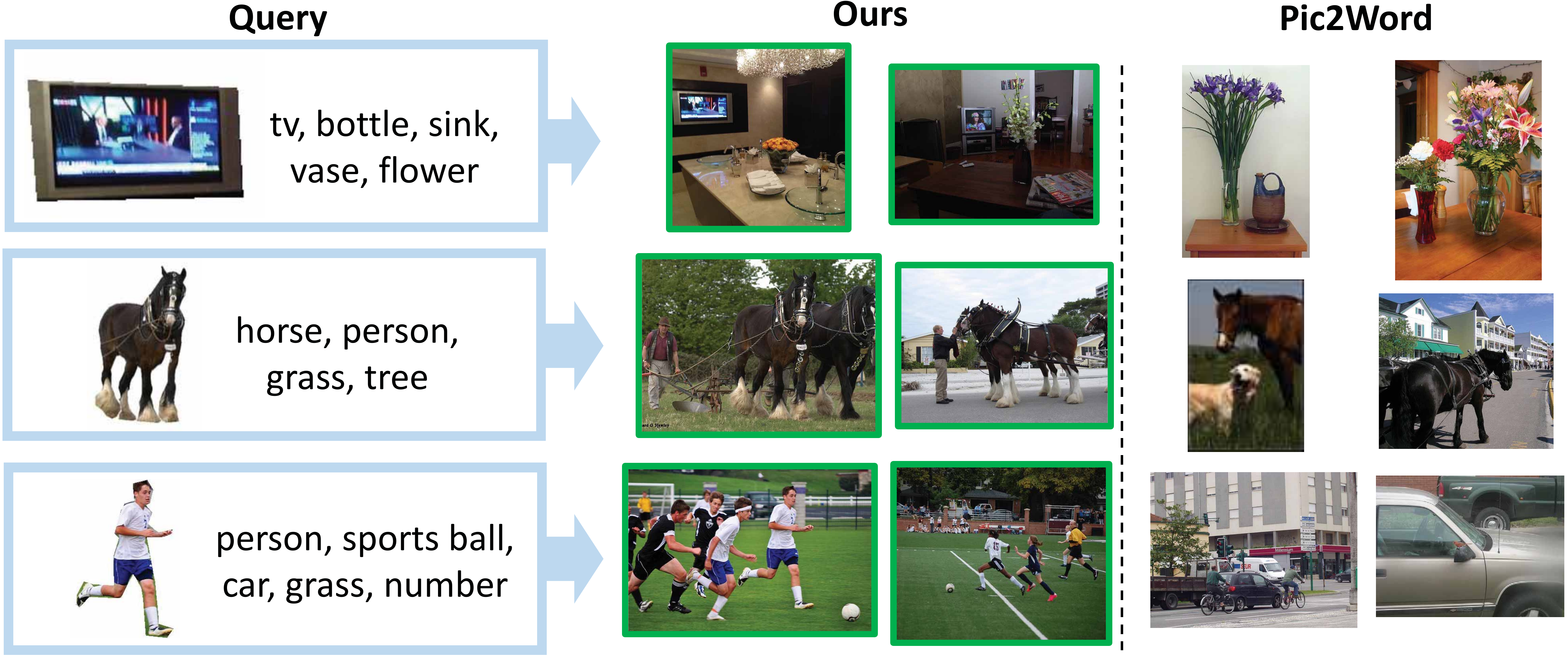}
    \caption{Retrieved results on the object composition task.}
    \label{fig:coco}
\end{figure}

\begin{figure}[t]
    \centering
    \centering
    \includegraphics[width=1.0\linewidth]{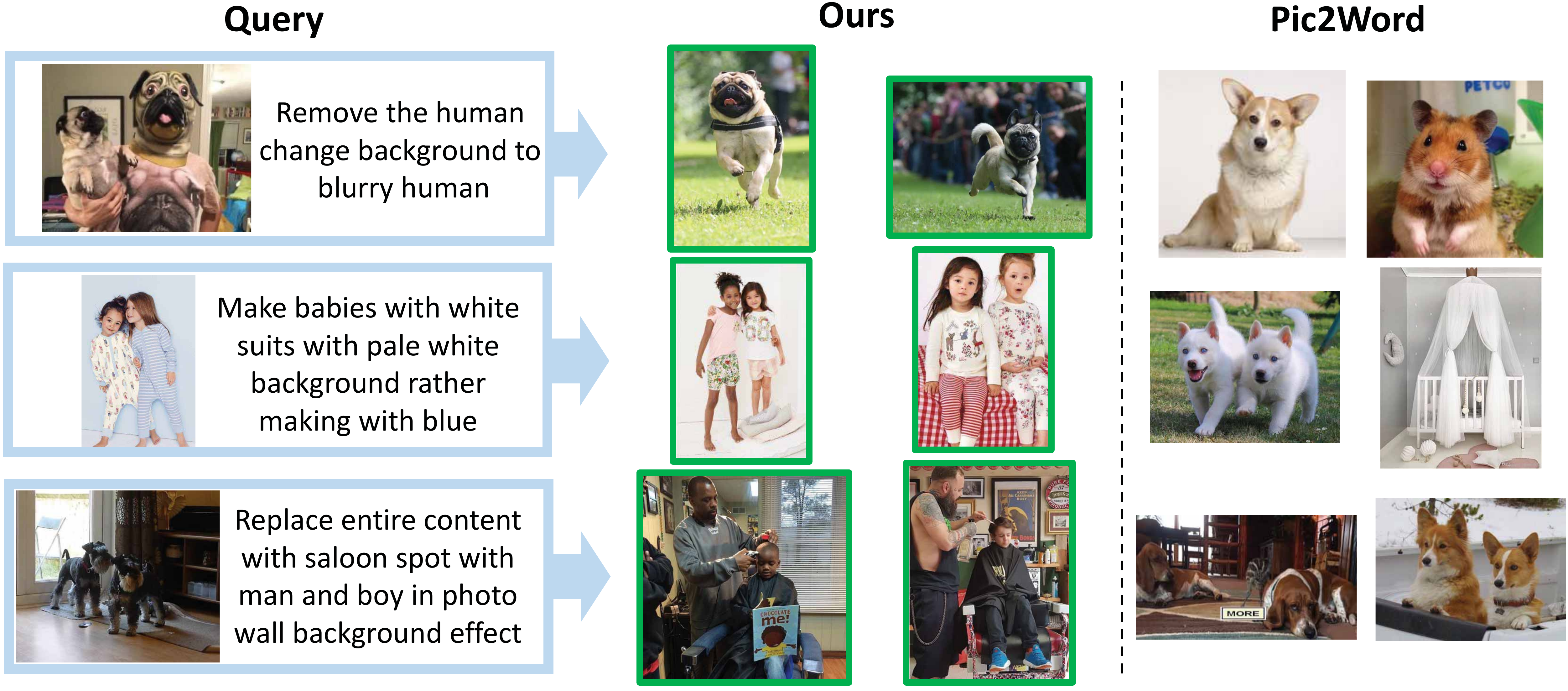}
    \caption{Retrieved results on the object manipulation task.}
    \label{fig:cirr}
\end{figure}

\begin{figure}[t]
    \centering
    \centering
    \includegraphics[width=1.0\linewidth]{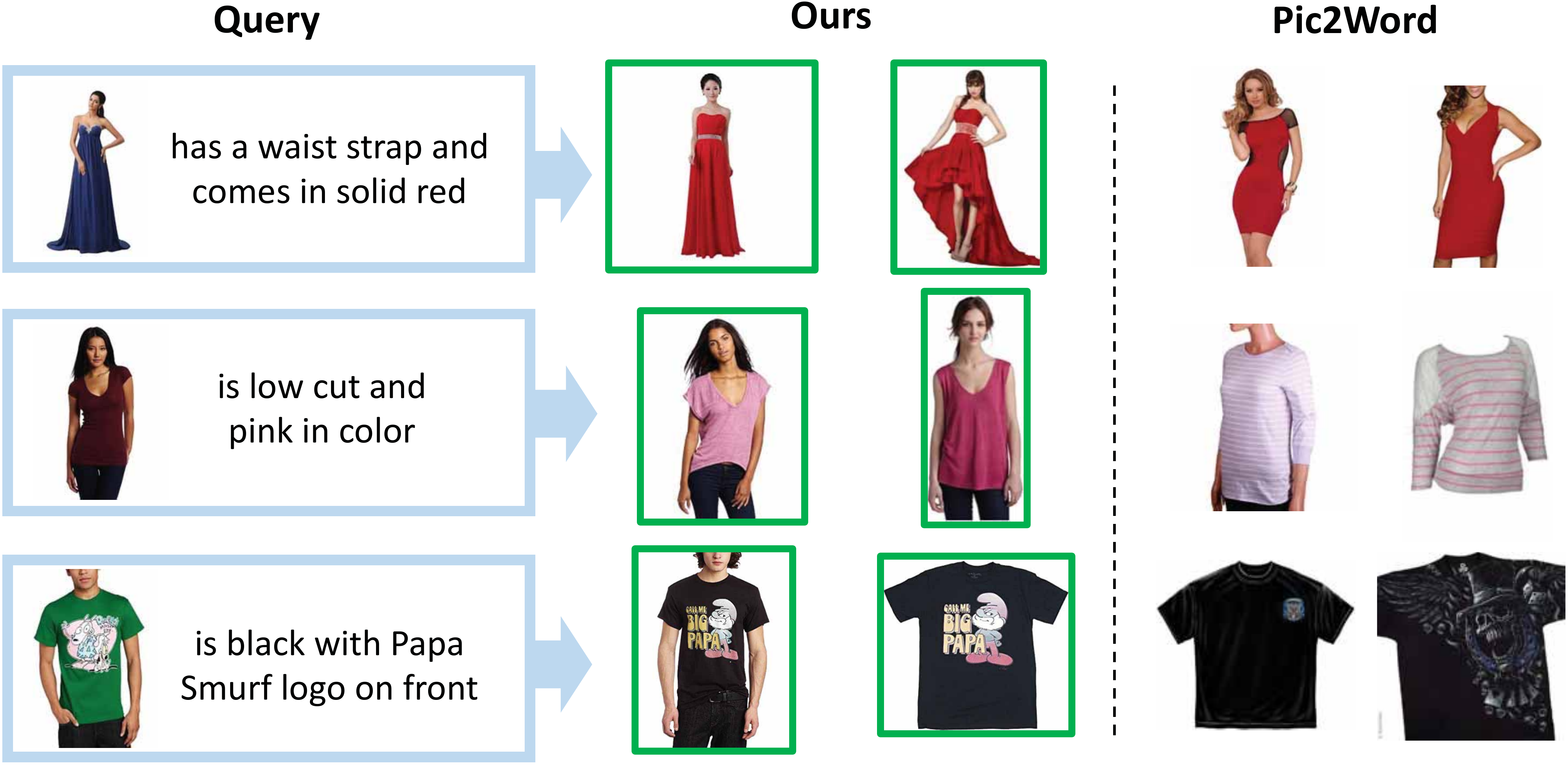}
    \caption{Retrieved results of attribute manipulation.}
    \vspace{-5pt}
    \label{fig:fashion}
\end{figure}

Table \ref{tab:imgnet}-\ref{tab:fashion} present the quantitative results. Figure \ref{fig:imgnet}-\ref{fig:fashion} show the corresponding qualitative results of our model and the most recent work Pic2Word. In the domain conversion results (Table \ref{tab:imgnet}), Context-I2W consistently outperforms existing approaches, including the supervised ones, and remarkably outperforms the State-of-the-Art (SoTA) Pic2Word by 3.60\% on average. As exemplified in Figure \ref{fig:imgnet}, Context-I2W accurately selects the targets (\textit{e.g.,}  train, cake, squirrel, and piano) for domain conversion. Pic2Word fails to focus on the correct visual parts because it coarsely maps the global image features without object selection ability, which is compensated by our specially designed VTE. 

In the experiments of object composition (Table \ref{tab:coco}), Context-I2W also outperforms the SoTA ZS-CIR model by 3.47\% on average while surpassing the supervised methods on all the metrics. This further proves that context-I2W enables the model to accurately convert the target visual content in the language token space to achieve cross-modal object combination, as depicted in Figure \ref{fig:coco}.

\begin{table*}[t]
\centering
\scalebox{1.05}
{\scriptsize
\begin{tabular}{cccccccccccc}
\toprule
                            &                                                          & \multicolumn{2}{c}{Dress}                                                               & \multicolumn{2}{c}{Shrit}                                                      & \multicolumn{2}{c}{TopTee}                                                              & \multicolumn{2}{c}{Average}                               \\ \cmidrule(lr){3-4}\cmidrule(lr){5-6}\cmidrule(lr){7-8}\cmidrule(lr){9-10}
                            %\cline{3-10} 
Supervision                 & Methods                                                  & R10                                  & R50                                              & R10                         & R50                                              & R10                                  & R50                                              & R10                         & R50                         \\ \hline
                            & Image-only                                               & 5.4                                  & \multicolumn{1}{c|}{13.9}                        & 9.9                         & \multicolumn{1}{c|}{20.8}                        & 8.3                                  & \multicolumn{1}{c|}{17.7}                        & 7.9                         & 17.5                        \\
                            & Text-only                                                & 13.6                                 & \multicolumn{1}{c|}{29.7}                        & 18.9                        & \multicolumn{1}{c|}{31.8}                        & 19.3                                 & \multicolumn{1}{c|}{37.0}                        & 17.3                        & 32.9                        \\
                            & Image+Text                                               & 16.3                                 & \multicolumn{1}{c|}{33.6}                        & 21.0                        & \multicolumn{1}{c|}{34.5}                        & 22.2                                 & \multicolumn{1}{c|}{39.0}                        & 19.8                        & 35.7                        \\
                            & Pic2Word (CVPR 2023)                                                & 20.0                                 & \multicolumn{1}{c|}{40.2}                        & 26.2                        & \multicolumn{1}{c|}{43.6}                        & 27.9                                 & \multicolumn{1}{c|}{47.4}                        & 24.7                        & 43.7                        \\
                            & SEARLE-XL (ICCV 2023)                                               & 20.3                                 & \multicolumn{1}{c|}{43.2}                        & 27.4                                 & \multicolumn{1}{c|}{45.7}                        & 29.3                                 & \multicolumn{1}{c|}{50.2}                        & 25.7                        & 46.3                        \\
                            \multirow{-6}{*}{ZERO-SHOT} & \textbf{Context-I2W}                                             & \textbf{23.1}                        & \multicolumn{1}{c|}{\textbf{45.3}}               & \textbf{29.7}                        & \multicolumn{1}{c|}{\textbf{48.6}}               & \textbf{30.6}                        & \multicolumn{1}{c|}{\textbf{52.9}}               & \textbf{27.8}               & \textbf{48.9}               \\ \cmidrule(lr){1-10}
CIRR                        & Combiner (CVPR 2022)                                                & 17.2                                 & \multicolumn{1}{c|}{37.9}                        & 23.7                        & \multicolumn{1}{c|}{39.4}                        & 24.1                                 & \multicolumn{1}{c|}{43.9}                        & 21.7                        & 40.4                        \\
Fashion-IQ                  & Combiner (CVPR 2022)                                                & 30.3                                 & \multicolumn{1}{c|}{54.5}                        & 37.2                        & \multicolumn{1}{c|}{55.8}                        & 39.2                                 & \multicolumn{1}{c|}{61.3}                        & 35.6                        & 57.2                        \\ 
Fashion-IQ                  & Combiner$^*$  (CVPR 2022)                                               & 31.6                                 & \multicolumn{1}{c|}{56.7}                        & 36.4                        & \multicolumn{1}{c|}{58.0}                        & 38.2                                 & \multicolumn{1}{c|}{62.4}                        & 35.4                        & 59.0                        \\
Fashion-IQ                  & CIRPLANT (ICCV 2021)                                              & 17.5                                 & \multicolumn{1}{c|}{40.4}                        & 17.5                        & \multicolumn{1}{c|}{38.8}                        & 21.6                                 & \multicolumn{1}{c|}{45.4}                        & 18.9                        & 41.5                        \\
Fashion-IQ                  & ARTEMIS  (ICLR 2022)                                                & 27.2                                 & \multicolumn{1}{c|}{52.4}                        & 21.8                        & \multicolumn{1}{c|}{43.6}                        & 29.2                                 & \multicolumn{1}{c|}{54.8}                        & 26.1                        & 50.3                        \\
Fashion-IQ                  & MAAF (arXiv 2020)                                                    & 23.8                                 & \multicolumn{1}{c|}{48.6}                        & 21.3                        & \multicolumn{1}{c|}{44.2}                        & 27.9                                 & \multicolumn{1}{c|}{53.6}                        & 24.3                        & 48.8                        \\ \bottomrule
\end{tabular}}
\caption{Results on Fashion-IQ for attribute manipulation.}
\label{tab:fashion}
\vspace{-10pt}
\end{table*}

We also evaluate the ability of foreground/background differentiation and fine-grained image editing through the object/scene manipulation task (Table \ref{tab:CIRR}). Context-I2W consistently outperforms existing ZS-CIR models and achieves a boost of 1.88\% on average over the best model. This is contributed to Context-I2W of selecting context-relevant visual information before mapping, effectively mitigating the impact of dataset-specific bias on ZS-CIR models \cite{Saito_2023_CVPR}. In Figure \ref{fig:cirr}, Context-I2W adaptively and accurately removes the specific object (row 1), changes the background (row 2), and even replaces the entire content of the image (row 3) based on the different context of descriptions.

The attribute manipulation task requires accurately localizing specific attributes in the entire image. As depicted in Table \ref{tab:fashion}, Context-I2W achieves an obvious average improvement of 2.35\% over the SoTA. Figure \ref{fig:fashion} provides further proves Context-I2W adeptly selects view representations of attributes and intricate visual details for word mapping.

\begin{figure}[t]
    \centering
    \includegraphics[width=1.0\linewidth]{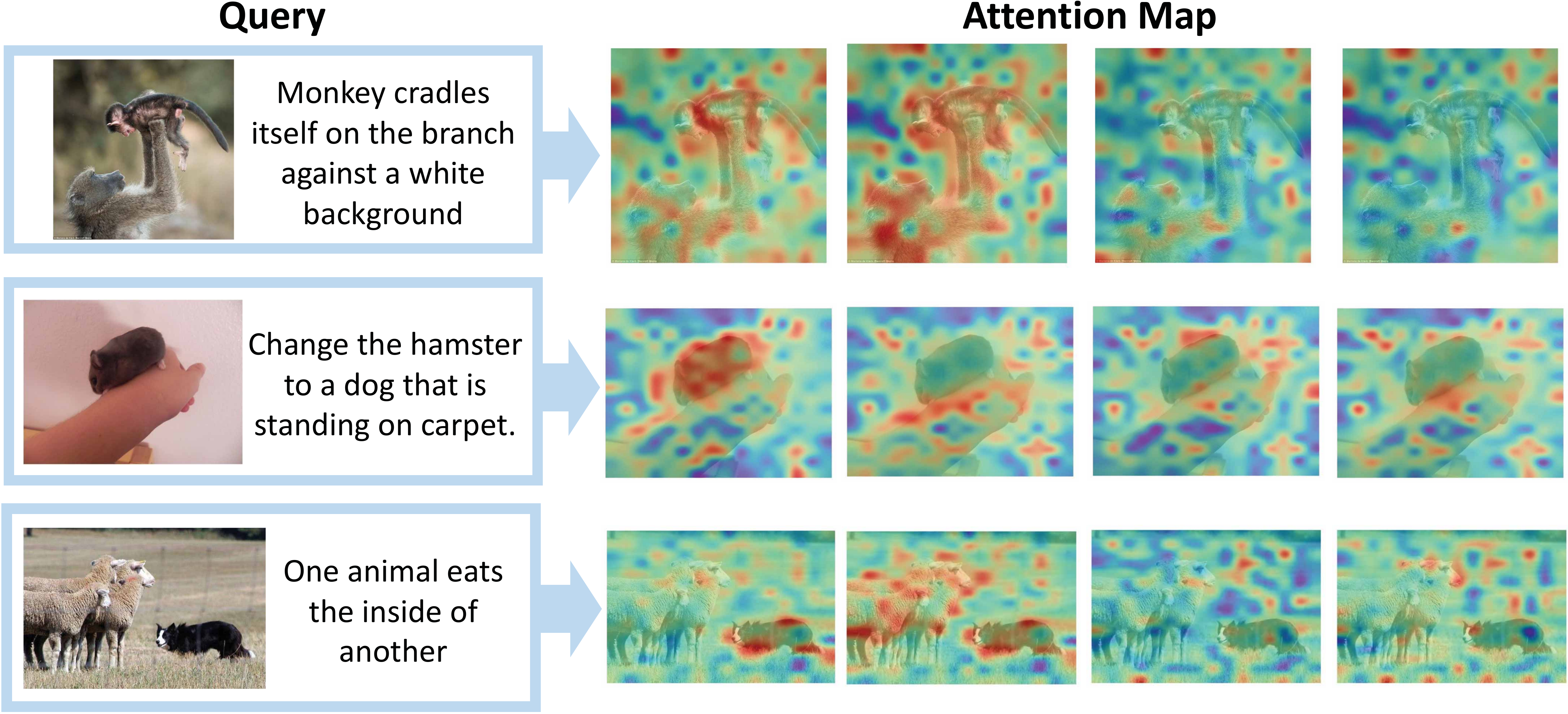}
    \caption{Attention map visualization of learnable queries.}
    \label{fig:attn}
    \vspace{-5pt}
\end{figure}

\subsection{Ablation Study}
\begin{table}[t]
\centering
\scalebox{1.05}
{\scriptsize
\begin{tabular}{llcclcc}
\toprule
\multicolumn{2}{l}{}   & \multicolumn{3}{c}{CIRR}                         & \multicolumn{2}{c}{Fashion-IQ} \\ \cmidrule(lr){3-5}\cmidrule(lr){6-7}
   & Methods           & R1   & R5   & R10                                & R10            & R50           \\ \hline
1. & full model        & 25.6 & 55.1 & \multicolumn{1}{l|}{68.5}          & 27.8           & 48.9          \\
2. & w/o context       & 21.4 & 50.3 & \multicolumn{1}{l|}{63.6}          & 23.8           & 42.5          \\
3. & w/o \texttt{[REPLACE]} & 24.4 & 53.3 & \multicolumn{1}{l|}{66.3}          & 26.4           & 45.3          \\
4. & w/o View Selector             & 24.0 & 53.5 & \multicolumn{1}{l|}{67.2}          & 25.7           & 46.1          \\
5. & w/o Target Extractor       & 23.3 & 51.5 & \multicolumn{1}{c|}{65.7}          & 26.1           & 44.8          \\
6. & w/o gate          & 24.5 & 54.3 & \multicolumn{1}{c|}{68.4}          & 25.9           & 45.4          \\ 
7. &  cross-attention   & 17.0 & 41.7 & \multicolumn{1}{l|}{55.0}          & 20.8           & 39.9          \\
8. & replace all noun  & 24.7 & 53.7 & \multicolumn{1}{l|}{67.1}          & 26.6           & 46.5          \\ 
9. & max pooling  & 24.3 & 54.1 & \multicolumn{1}{c|}{67.4}          & 25.8           & 46.2          \\
10. & directly input 4 tokens & 23.7 & 52.8 & \multicolumn{1}{c|}{66.2}          & 25.5           & 45.7          \\ 
\bottomrule
\end{tabular}}
\caption{Ablation study on CIRR and FashionIQ.}
\vspace{-5pt}
\label{tab:ablation}
\end{table}

In table \ref{tab:ablation}, we evaluate the contribution of the core components in Context-I2W on CIRR and FashionIQ. \textbf{(1) In models `2-3', we evaluate the importance of context encoding approach.} Removing the context embedding  $\boldsymbol{t}$ (model '2') results in a significant drop of 4.86\% on average. When subject replacement is not applied (model `3'), the performance declined by an average of 2.04\%, indicating that \texttt{[REPLACE]} is beneficial for extracting target-relevant visual information. \textbf{(2) In models `4-6', we assess the importance of key modules in image-to-word mapping process.} Removing either IVS (model `4') or VTE (model `5') causes obvious performance decrease. Notably, model `5' brings a bigger decrease since the two tasks require manipulation of detailed visual parts, which is effectively captured by our extractor module. By directly summing the local and global features instead of using the gating strategy (model '6'), the performance drops by 1.48\%. It indicates the necessity to capture complementary information from the two sources adaptively. \textbf{(3) Models `7-8' evaluate the effect of alternative solutions for key modules.} In model `7', we  replace the Context-I2W with a typical cross-attention network ($Q=\text{description}, K=V=\text{image}$). The results drop significantly by 10.30\% on average, confirming the effectiveness of the Context-I2W mapping strategy. In model '8', we replaced all nouns in the captions with \texttt{[REPLACE]}, which disrupts the relevant information about the target and results in an average drop of 1.46\%.  \textbf{(4) Models `9-10' evaluate the effect of other pooling strategies.} We compared the performance of average pooling (model `1') with max pooling (model `2') and directly input four pseudo-tokens (model `3'). The results indicate that average pooling outperforms these alternatives. Specifically, max pooling results in an average performance drop of 1.62\% as it undermines the diverse views captured by the learnable queries. Similarly, direct input of four distinct pseudo-tokens that are individual with each other results in a substantial performance drop of 2.40\% because it disrupts the syntactic structure of text input of the CLIP Language Encoder. 

\subsection{Interpretability of Learnable Query}
In Figure \ref{fig:attn}, we visualize the attention maps of each learnable query from the last block. The four queries obviously attend to different targets in the whole image: the first two queries mainly focus on the object attributes and foreground information, while the last two queries mostly consider the background and scene information. This pattern is usually consistent (Please refer to the supplementary materials for more samples). These attention maps provide insight that our model is interpretable in extracting specific visual features of different targets, which supports fine-grained visual information selection for the context.

\subsection{Effectiveness and efficiency Analysis} 
Our approach obtains significant improvement over four widely compared ZR-CIR tasks from 1.88\% to 3.60\% compared to the SoTA models. Since we design dedicated modules for accurate mapping, our model size  (65.3M) is bigger than the simple 3-layer MLP mapping (0.9M) of Pic2Word. Accordingly, our training time (24h) increased by 8 hours compared to Pic2Word in the same settings. It’s worth noting that our model using only 50\% of the pre-training data achieves comparable performance to  SoTA models (results are shown in Appendix C), which makes up for training time. Our inference time (0.026s) is slightly 0.007s slower than Pic2Word. % Designing lightweight as well as accurate models will be our future work. 

\begin{figure}[t]
    \centering
    \centering
    \includegraphics[width=1.0\linewidth]{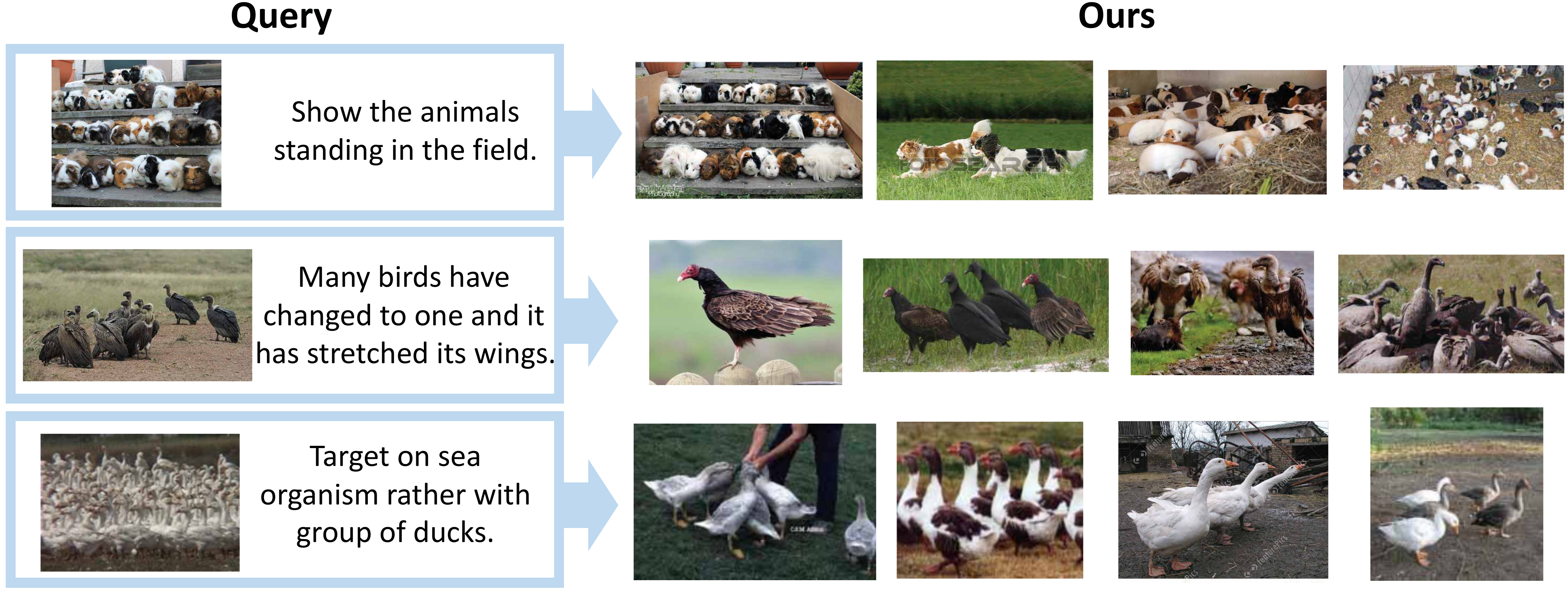}
    \caption{Visualization on common failure cases.}
    \label{fig:failure}
\end{figure}

\subsection{Discussion on common failure cases} 
In Figure \ref{fig:failure}, we visualize several common failure cases of our Context-I2W. Notably, our method struggles in scenarios where the image comprises numerous small objects. We believe this is because the learned queries find it challenging to capture the fine-grained features of small objects within the patch-level local information provided by CLIP. 

\section{Conclusion}
In this paper,  we propose a novel image-to-word mapping network that leverages manipulation descriptions and learnable queries for context-dependent visual information selection and accurate mapping. Context-I2W  shows strong generalization ability and remarkably improves the best performance of existing approaches on four diverse ZS-CIR tasks. It inspires the vision-to-language alignment mechanism and impacts diverse vision and language applications. How to design more lightweight models with high retrieval performance will be the future work.

\bibliography{aaai24}

\begin{thebibliography}{38}
\providecommand{\natexlab}[1]{#1}

\bibitem[{Alayrac et~al.(2022)Alayrac, Donahue, Luc, Miech, Barr, Hasson, Lenc, Mensch, Millican, Reynolds, Ring, Rutherford, Cabi, Han, Gong, Samangooei, Monteiro, Menick, Borgeaud, Brock, Nematzadeh, Sharifzadeh, Bi\'{n}kowski, Barreira, Vinyals, Zisserman, and Simonyan}]{NEURIPS2022_960a172b}
Alayrac, J.-B.; Donahue, J.; Luc, P.; Miech, A.; Barr, I.; Hasson, Y.; Lenc, K.; Mensch, A.; Millican, K.; Reynolds, M.; Ring, R.; Rutherford, E.; Cabi, S.; Han, T.; Gong, Z.; Samangooei, S.; Monteiro, M.; Menick, J.~L.; Borgeaud, S.; Brock, A.; Nematzadeh, A.; Sharifzadeh, S.; Bi\'{n}kowski, M.~a.; Barreira, R.; Vinyals, O.; Zisserman, A.; and Simonyan, K. 2022.
\newblock Flamingo: a Visual Language Model for Few-Shot Learning.
\newblock In Koyejo, S.; Mohamed, S.; Agarwal, A.; Belgrave, D.; Cho, K.; and Oh, A., eds., \emph{Advances in Neural Information Processing Systems}, volume~35, 23716--23736.

\bibitem[{Bachlechner et~al.(2021)Bachlechner, Majumder, Mao, Cottrell, and McAuley}]{bachlechner2021rezero}
Bachlechner, T.; Majumder, B.~P.; Mao, H.; Cottrell, G.; and McAuley, J. 2021.
\newblock Rezero is all you need: Fast convergence at large depth.
\newblock In \emph{Uncertainty in Artificial Intelligence}, 1352--1361.

\bibitem[{Baldrati et~al.(2023)Baldrati, Agnolucci, Bertini, and Del~Bimbo}]{baldrati2023zero}
Baldrati, A.; Agnolucci, L.; Bertini, M.; and Del~Bimbo, A. 2023.
\newblock Zero-Shot Composed Image Retrieval with Textual Inversion.
\newblock \emph{arXiv:2303.15247}.

\bibitem[{Baldrati et~al.(2022)Baldrati, Bertini, Uricchio, and Del~Bimbo}]{Baldrati_2022_CVPR}
Baldrati, A.; Bertini, M.; Uricchio, T.; and Del~Bimbo, A. 2022.
\newblock Effective Conditioned and Composed Image Retrieval Combining CLIP-Based Features.
\newblock In \emph{Proceedings of the IEEE/CVF Conference on Computer Vision and Pattern Recognition}, 21466--21474.

\bibitem[{Brown et~al.(2020)Brown, Mann, Ryder, Subbiah, Kaplan, Dhariwal, Neelakantan, Shyam, Sastry, Askell, Agarwal, Herbert-Voss, Krueger, Henighan, Child, Ramesh, Ziegler, Wu, Winter, Hesse, Chen, Sigler, Litwin, Gray, Chess, Clark, Berner, McCandlish, Radford, Sutskever, and Amodei}]{brown2020language}
Brown, T.; Mann, B.; Ryder, N.; Subbiah, M.; Kaplan, J.~D.; Dhariwal, P.; Neelakantan, A.; Shyam, P.; Sastry, G.; Askell, A.; Agarwal, S.; Herbert-Voss, A.; Krueger, G.; Henighan, T.; Child, R.; Ramesh, A.; Ziegler, D.; Wu, J.; Winter, C.; Hesse, C.; Chen, M.; Sigler, E.; Litwin, M.; Gray, S.; Chess, B.; Clark, J.; Berner, C.; McCandlish, S.; Radford, A.; Sutskever, I.; and Amodei, D. 2020.
\newblock Language Models are Few-Shot Learners.
\newblock In Larochelle, H.; Ranzato, M.; Hadsell, R.; Balcan, M.; and Lin, H., eds., \emph{Advances in Neural Information Processing Systems}, volume~33, 1877--1901. Curran Associates, Inc.

\bibitem[{Carion et~al.(2020)Carion, Massa, Synnaeve, Usunier, Kirillov, and Zagoruyko}]{carion2020end}
Carion, N.; Massa, F.; Synnaeve, G.; Usunier, N.; Kirillov, A.; and Zagoruyko, S. 2020.
\newblock End-to-end object detection with transformers.
\newblock In \emph{European conference on computer vision}, 213--229.

\bibitem[{Chen, Gong, and Bazzani(2020)}]{Chen_2020_CVPR}
Chen, Y.; Gong, S.; and Bazzani, L. 2020.
\newblock Image Search With Text Feedback by Visiolinguistic Attention Learning.
\newblock In \emph{Proceedings of the IEEE/CVF Conference on Computer Vision and Pattern Recognition}, 3001--3011.

\bibitem[{Cohen et~al.(2022)Cohen, Gal, Meirom, Chechik, and Atzmon}]{10.1007/978-3-031-20044-1_32}
Cohen, N.; Gal, R.; Meirom, E.~A.; Chechik, G.; and Atzmon, Y. 2022.
\newblock ``This Is My Unicorn, Fluffy'': Personalizing Frozen Vision-Language Representations.
\newblock In \emph{European conference on computer vision}, 558--577.

\bibitem[{Datta et~al.(2008)Datta, Joshi, Li, and Wang}]{datta2008image}
Datta, R.; Joshi, D.; Li, J.; and Wang, J.~Z. 2008.
\newblock Image retrieval: Ideas, influences, and trends of the new age.
\newblock \emph{ACM Computing Surveys}, 40(2): 1--60.

\bibitem[{Delmas et~al.(2022)Delmas, Rezende, Csurka, and Larlus}]{delmas2022artemis}
Delmas, G.; Rezende, R.~S.; Csurka, G.; and Larlus, D. 2022.
\newblock {ARTEMIS}: Attention-based Retrieval with Text-Explicit Matching and Implicit Similarity.
\newblock In \emph{International Conference on Learning Representations}.

\bibitem[{Deng et~al.(2009)Deng, Dong, Socher, Li, Li, and Fei-Fei}]{deng2009imagenet}
Deng, J.; Dong, W.; Socher, R.; Li, L.-J.; Li, K.; and Fei-Fei, L. 2009.
\newblock Imagenet: A large-scale hierarchical image database.
\newblock In \emph{Computer Vision and Pattern Recognition}, 248--255.

\bibitem[{Dodds et~al.(2020)Dodds, Culpepper, Herdade, Zhang, and Boakye}]{dodds2020modality}
Dodds, E.; Culpepper, J.; Herdade, S.; Zhang, Y.; and Boakye, K. 2020.
\newblock Modality-agnostic attention fusion for visual search with text feedback.
\newblock \emph{arXiv:2007.00145}.

\bibitem[{Goenka et~al.(2022)Goenka, Zheng, Jaiswal, Chada, Wu, Hedau, and Natarajan}]{Goenka_2022_CVPR}
Goenka, S.; Zheng, Z.; Jaiswal, A.; Chada, R.; Wu, Y.; Hedau, V.; and Natarajan, P. 2022.
\newblock FashionVLP: Vision Language Transformer for Fashion Retrieval With Feedback.
\newblock In \emph{Proceedings of the IEEE/CVF Conference on Computer Vision and Pattern Recognition}, 14105--14115.

\bibitem[{Han et~al.(2022)Han, Yu, Zhu, Zhang, Song, and Xiang}]{10.1007/978-3-031-19833-5_37}
Han, X.; Yu, L.; Zhu, X.; Zhang, L.; Song, Y.-Z.; and Xiang, T. 2022.
\newblock FashionViL: Fashion-Focused Vision-and-Language Representation Learning.
\newblock In \emph{European conference on computer vision}, 634--651.

\bibitem[{Hendrycks et~al.(2021)Hendrycks, Basart, Mu, Kadavath, Wang, Dorundo, Desai, Zhu, Parajuli, Guo, Song, Steinhardt, and Gilmer}]{Hendrycks_2021_ICCV}
Hendrycks, D.; Basart, S.; Mu, N.; Kadavath, S.; Wang, F.; Dorundo, E.; Desai, R.; Zhu, T.; Parajuli, S.; Guo, M.; Song, D.; Steinhardt, J.; and Gilmer, J. 2021.
\newblock The Many Faces of Robustness: A Critical Analysis of Out-of-Distribution Generalization.
\newblock In \emph{Proceedings of the IEEE/CVF International Conference on Computer Vision}, 8340--8349.

\bibitem[{Hochreiter and Schmidhuber(1997)}]{hochreiter1997long}
Hochreiter, S.; and Schmidhuber, J. 1997.
\newblock Long short-term memory.
\newblock \emph{Neural computation}, 9(8): 1735--1780.

\bibitem[{Honnibal et~al.(2020)Honnibal, Montani, Van~Landeghem, Boyd et~al.}]{honnibal2020spacy}
Honnibal, M.; Montani, I.; Van~Landeghem, S.; Boyd, A.; et~al. 2020.
\newblock spaCy: Industrial-strength natural language processing in python.

\bibitem[{Kumari et~al.(2023)Kumari, Zhang, Zhang, Shechtman, and Zhu}]{Kumari_2023_CVPR}
Kumari, N.; Zhang, B.; Zhang, R.; Shechtman, E.; and Zhu, J.-Y. 2023.
\newblock Multi-Concept Customization of Text-to-Image Diffusion.
\newblock In \emph{Proceedings of the IEEE/CVF Conference on Computer Vision and Pattern Recognition}, 1931--1941.

\bibitem[{Li et~al.(2023)Li, Li, Savarese, and Hoi}]{li2023blip2}
Li, J.; Li, D.; Savarese, S.; and Hoi, S. 2023.
\newblock BLIP-2: Bootstrapping Language-Image Pre-training with Frozen Image Encoders and Large Language Models.
\newblock arXiv:2301.12597.

\bibitem[{Li et~al.(2022)Li, Li, Xiong, and Hoi}]{pmlr-v162-li22n}
Li, J.; Li, D.; Xiong, C.; and Hoi, S. 2022.
\newblock {BLIP}: Bootstrapping Language-Image Pre-training for Unified Vision-Language Understanding and Generation.
\newblock In \emph{Proceedings of the 39th International Conference on Machine Learning}, 12888--12900.

\bibitem[{Li et~al.(2021)Li, Selvaraju, Gotmare, Joty, Xiong, and Hoi}]{NEURIPS2021_50525975}
Li, J.; Selvaraju, R.; Gotmare, A.; Joty, S.; Xiong, C.; and Hoi, S. C.~H. 2021.
\newblock Align before Fuse: Vision and Language Representation Learning with Momentum Distillation.
\newblock In \emph{Advances in Neural Information Processing Systems}, 9694--9705.

\bibitem[{Li et~al.(2020)Li, Yin, Li, Zhang, Hu, Zhang, Wang, Hu, Dong, Wei et~al.}]{li2020oscar}
Li, X.; Yin, X.; Li, C.; Zhang, P.; Hu, X.; Zhang, L.; Wang, L.; Hu, H.; Dong, L.; Wei, F.; et~al. 2020.
\newblock Oscar: Object-semantics aligned pre-training for vision-language tasks.
\newblock In \emph{European Conference on Computer Vision}, 121--137.

\bibitem[{Lin et~al.(2014)Lin, Maire, Belongie, Hays, Perona, Ramanan, Doll{\'a}r, and Zitnick}]{10.1007/978-3-319-10602-1_48}
Lin, T.-Y.; Maire, M.; Belongie, S.; Hays, J.; Perona, P.; Ramanan, D.; Doll{\'a}r, P.; and Zitnick, C.~L. 2014.
\newblock Microsoft COCO: Common Objects in Context.
\newblock In Fleet, D.; Pajdla, T.; Schiele, B.; and Tuytelaars, T., eds., \emph{European Conference on Computer Vision}, 740--755.

\bibitem[{Liu et~al.(2021)Liu, Rodriguez-Opazo, Teney, and Gould}]{Liu_2021_ICCV}
Liu, Z.; Rodriguez-Opazo, C.; Teney, D.; and Gould, S. 2021.
\newblock Image Retrieval on Real-Life Images With Pre-Trained Vision-and-Language Models.
\newblock In \emph{Proceedings of the IEEE/CVF International Conference on Computer Vision}, 2125--2134.

\bibitem[{Loshchilov and Hutter(2018)}]{loshchilov2018decoupled}
Loshchilov, I.; and Hutter, F. 2018.
\newblock Decoupled Weight Decay Regularization.
\newblock In \emph{International Conference on Learning Representations}.

\bibitem[{Mokady, Hertz, and Bermano(2021)}]{mokady2021clipcap}
Mokady, R.; Hertz, A.; and Bermano, A.~H. 2021.
\newblock ClipCap: CLIP Prefix for Image Captioning.
\newblock arXiv:2111.09734.

\bibitem[{Radford et~al.(2021)Radford, Kim, Hallacy, Ramesh, Goh, Agarwal, Sastry, Askell, Mishkin, Clark, Krueger, and Sutskever}]{radford2021learning}
Radford, A.; Kim, J.~W.; Hallacy, C.; Ramesh, A.; Goh, G.; Agarwal, S.; Sastry, G.; Askell, A.; Mishkin, P.; Clark, J.; Krueger, G.; and Sutskever, I. 2021.
\newblock Learning Transferable Visual Models From Natural Language Supervision.
\newblock In \emph{Proceedings of the International Conference on Machine Learning}, 8748--8763.

\bibitem[{Saito et~al.(2023)Saito, Sohn, Zhang, Li, Lee, Saenko, and Pfister}]{Saito_2023_CVPR}
Saito, K.; Sohn, K.; Zhang, X.; Li, C.-L.; Lee, C.-Y.; Saenko, K.; and Pfister, T. 2023.
\newblock Pic2Word: Mapping Pictures to Words for Zero-Shot Composed Image Retrieval.
\newblock In \emph{Proceedings of the IEEE/CVF Conference on Computer Vision and Pattern Recognition}, 19305--19314.

\bibitem[{Sharma et~al.(2018)Sharma, Ding, Goodman, and Soricut}]{DBLP:conf/acl/SoricutDSG18}
Sharma, P.; Ding, N.; Goodman, S.; and Soricut, R. 2018.
\newblock Conceptual Captions: A Cleaned, Hypernymed, Image Alt-text Dataset For Automatic Image Captioning.
\newblock In \emph{Annual Meeting of the Association for Computational Linguistics}, 2556--2565.

\bibitem[{Shi et~al.(2023)Shi, Zhang, Yin, Xie, Zhang, Fan, Shi, and Qu}]{Shi_2023_ICCV}
Shi, J.; Zhang, Y.; Yin, X.; Xie, Y.; Zhang, Z.; Fan, J.; Shi, Z.; and Qu, Y. 2023.
\newblock Dual Pseudo-Labels Interactive Self-Training for Semi-Supervised Visible-Infrared Person Re-Identification.
\newblock In \emph{Proceedings of the IEEE/CVF International Conference on Computer Vision (ICCV)}, 11218--11228.

\bibitem[{Song et~al.(2022)Song, Dong, Zhang, Liu, and Wei}]{song2022clip}
Song, H.; Dong, L.; Zhang, W.-N.; Liu, T.; and Wei, F. 2022.
\newblock CLIP Models are Few-shot Learners: Empirical Studies on VQA and Visual Entailment.
\newblock arXiv:2203.07190.

\bibitem[{Tam, Raffel, and Bansal(2023)}]{tam2023simple}
Tam, D.; Raffel, C.; and Bansal, M. 2023.
\newblock Simple Weakly-Supervised Image Captioning via {CLIP}'s Multimodal Embeddings.
\newblock In \emph{The AAAI-23 Workshop on Creative AI Across Modalities}.

\bibitem[{Vo et~al.(2019{\natexlab{a}})Vo, Jiang, Sun, Murphy, Li, Fei-Fei, and Hays}]{Vo_2019_CVPR}
Vo, N.; Jiang, L.; Sun, C.; Murphy, K.; Li, L.-J.; Fei-Fei, L.; and Hays, J. 2019{\natexlab{a}}.
\newblock Composing Text and Image for Image Retrieval - an Empirical Odyssey.
\newblock In \emph{Proceedings of the IEEE/CVF Conference on Computer Vision and Pattern Recognition}, 6439--6448.

\bibitem[{Vo et~al.(2019{\natexlab{b}})Vo, Jiang, Sun, Murphy, Li, Fei-Fei, and Hays}]{vo2019composing}
Vo, N.; Jiang, L.; Sun, C.; Murphy, K.; Li, L.-J.; Fei-Fei, L.; and Hays, J. 2019{\natexlab{b}}.
\newblock Composing text and image for image retrieval-an empirical odyssey.
\newblock In \emph{Proceedings of the IEEE/CVF conference on computer vision and pattern recognition}, 6439--6448.

\bibitem[{Wu et~al.(2021)Wu, Gao, Guo, Al-Halah, Rennie, Grauman, and Feris}]{Wu_2021_CVPR}
Wu, H.; Gao, Y.; Guo, X.; Al-Halah, Z.; Rennie, S.; Grauman, K.; and Feris, R. 2021.
\newblock Fashion IQ: A New Dataset Towards Retrieving Images by Natural Language Feedback.
\newblock In \emph{Proceedings of the IEEE/CVF Conference on Computer Vision and Pattern Recognition}, 11307--11317.

\bibitem[{Zhang et~al.(2021)Zhang, Li, Hu, Yang, Zhang, Wang, Choi, and Gao}]{zhang2021vinvl}
Zhang, P.; Li, X.; Hu, X.; Yang, J.; Zhang, L.; Wang, L.; Choi, Y.; and Gao, J. 2021.
\newblock Vinvl: Revisiting visual representations in vision-language models.
\newblock In \emph{Proceedings of the IEEE/CVF Conference on Computer Vision and Pattern Recognition}, 5579--5588.

\bibitem[{Zhou et~al.(2022)Zhou, Yang, Loy, and Liu}]{Zhou_2022_CVPR}
Zhou, K.; Yang, J.; Loy, C.~C.; and Liu, Z. 2022.
\newblock Conditional Prompt Learning for Vision-Language Models.
\newblock In \emph{Proceedings of the IEEE/CVF Conference on Computer Vision and Pattern Recognition}, 16816--16825.

\bibitem[{Zhu et~al.(2023)Zhu, Yan, Lu, Xu, Wang, Eckstein, and Wang}]{zhu2023visualize}
Zhu, W.; Yan, A.; Lu, Y.; Xu, W.; Wang, X.~E.; Eckstein, M.; and Wang, W.~Y. 2023.
\newblock Visualize Before You Write: Imagination-Guided Open-Ended Text Generation.
\newblock arXiv:2210.03765.

\end{thebibliography}

\clearpage

\appendix
\section{Appendix}

This document presents supplementary materials supporting our main submission. In Section A, we provide an analysis of the number of learnable queries. In Section B, we provide the pseudo-code for the Visual Target Extractor. In Section C, we present additional evidence to establish the effectiveness and efficiency of Context-I2W. Section D showcases more samples of the attention maps generated by each learnable query, demonstrating that Context-I2W is interpretable in extracting specific visual features of different targets. Section E provides further implementation details for Context-I2W, including additional information about the datasets (Section E.1)

\subsection{A. Analysis of the number of learnable queries.}
\begin{figure}[th]
    \centering
    \centering
    \includegraphics[width=1.0\linewidth]{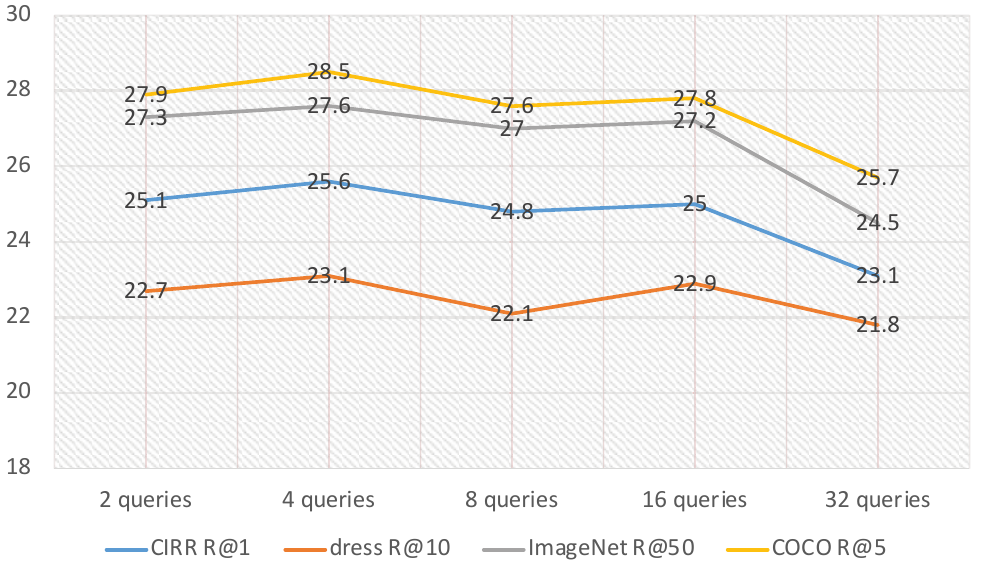}
    \caption{Analysis of the number of learnable queries.}
    \label{fig:num_ablation}
\end{figure}

We conduct analysis on the number of learnable query embedding $\boldsymbol{X}=\{\boldsymbol{x}_k\}_{k=1}^n\in{\mathbb{R}^{d\times{n}}}$ as shown in Figure \ref{fig:num_ablation}. We find that $n = 2$ results in not learning sufficient context-dependant visual details, but when $n$ is added to $32$, it is redundant and unhelpful. We finally choose $n = 4$, which gives the best result among different settings.

\subsection{B. Algorithm of Visual Target Extractor} The Visual Target Extractor is to eliminate context-independent visual features from $\tilde{\boldsymbol{v}}_i$, thereby facilitating context-dependant word mapping. Algorithm \ref{alg:algorithm_sampler} outlines the pseudo-code for the Visual Target Extractor. We create a fixed number of learnable embeddings as latent queries to achieve a fine-grained sampling of text-relevant visual features. The objective is to minimize the presence of context-independent visual features within $\tilde{\boldsymbol{v}}_i$. These learnable embeddings are then employed in a Transformer to execute cross-attention with the chosen local visual features $\{\tilde{\boldsymbol{v}}_i\}_{i=2}^m$. The number of output tokens produced by the Context-Dependent Sampler matches the count of learnable embeddings. To enhance interaction between learnable embeddings and local visual features, we concatenate the learnable embeddings with keys and values during the cross-attention process. Each learned query, interacting with context-dependent local visual features, captures diverse image regions from different angles based on the context, as shown in Figure 1 ((main paper)). This contrast is evident when comparing the example of the \textit{Target Extractor} to the \textit{Intent View Selector} in Figure 1(b)(main paper), showing that the learnable query effectively selects more fine-grained visual information. To achieve a dynamic ratio during the fusion of global and local features, allowing context-based acquisition of the pseudo token $S_*$, we utilize a tanh-gating mechanism \cite{hochreiter1997long}.

\section{C. More Effectiveness and Efficiency Analysis}
\label{more_effectiveness}
\begin{algorithm}[tb]
\caption{Visual Target Extractor’s process.}
\label{alg:algorithm_sampler}  
\textbf{Input}: batch of rotated image features $\tilde{\boldsymbol{V}}=  \{\tilde{\boldsymbol{v}}_i\}_{i=1}^m$, where $\tilde{\boldsymbol{v}}_1$ is the global feature $\tilde{\boldsymbol{v}}_g$, $N_{layer}$.\\
\textbf{Parameter}:  a set of learnable embeddings $\boldsymbol{X}\in{\mathbb{R}^{d\times{n}}}$
,  8-heads attention layer $Attn$, 3-layers FC layers $f_M$, $gate_\alpha$. \\
\textbf{Output}: pseudo token $S_*$
\begin{algorithmic}[1] %[1] enables line numbers
\STATE Initialize $\boldsymbol{X}\in{\mathbb{R}^{d\times{n}}}$, $Attn$, $f_M$ randomly.
\STATE Let $\boldsymbol{X}^i_{att}= \{\tilde{\boldsymbol{v}}_i\}_{i=2}^m$, $t=1$
\WHILE{$t \le N_{layer}$}
\STATE $\boldsymbol{X}^{i+1}_{att} = \boldsymbol{X}^{i}_{att} + Attn_t$(q=$\boldsymbol{q}$, k=$concat([\boldsymbol{X}^{i}_{att}, \boldsymbol{q}])$, v=$concat([\boldsymbol{X}^{i}_{att}, \boldsymbol{q}])$)
\STATE $\boldsymbol{X}^{i+1}_{att} = \boldsymbol{X}^{i+1}_{att} + f_{M_t}(\boldsymbol{X}^{i+1}_{att})$
\STATE $t = t + 1$
\ENDWHILE

$S_*$ = $f_{M_{g}}$($\tilde{\boldsymbol{v}}_g) + tanh(gate_\alpha) \cdot avg$($f_{M_{l}}(\boldsymbol{X}_{output}))$

\STATE \textbf{return} $S_*$
\end{algorithmic}
\end{algorithm}

\begin{table*}[t]
\centering
\scalebox{1.11}
{\scriptsize
\begin{tabular}{cccccccccccc}
\toprule
                           &              & \multicolumn{2}{c}{Cartoon}                       & \multicolumn{2}{c}{Origami}                        & \multicolumn{2}{c}{Toy}                            & \multicolumn{2}{c}{Sculpture}                      & \multicolumn{2}{c}{Average}   \\  \cmidrule(lr){3-4}\cmidrule(lr){5-6}\cmidrule(lr){7-8}\cmidrule(lr){9-10}\cmidrule(lr){11-12}
                           %\cline{3-10}
Supervision                & Methods      & R10          & R50                                & R10           & R50                                & R10           & R50                                & R10           & R50                                & R10           & R50           \\ \hline
\multirow{6}{*}{ZERO-SHOT} & Image-only   & 0.3          & \multicolumn{1}{c|}{4.5}           & 0.2           & \multicolumn{1}{c|}{1.8}           & 0.6           & \multicolumn{1}{c|}{5.7}           & 0.3           & \multicolumn{1}{c|}{4.0}           & 0.4           & 4.0           \\
                           & Text-only    & 0.2          & \multicolumn{1}{c|}{1.1}           & 0.8           & \multicolumn{1}{c|}{3.7}           & 0.8           & \multicolumn{1}{c|}{2.4}           & 0.4           & \multicolumn{1}{c|}{2.0}           & 0.5           & 2.3           \\
                           & Image+Text   & 2.2          & \multicolumn{1}{c|}{13.3}          & 2.0           & \multicolumn{1}{c|}{10.3}          & 1.2           & \multicolumn{1}{c|}{9.7}           & 1.6           & \multicolumn{1}{c|}{11.6}          & 1.7           & 11.2          \\
                           & Pic2Word    & 8.0          & \multicolumn{1}{c|}{21.9}          & 13.5          & \multicolumn{1}{c|}{25.6}          & 8.7           & \multicolumn{1}{c|}{21.6}          & 10.0          & \multicolumn{1}{c|}{23.8}          & 10.1          & 23.2          \\ 
                           & Context-I2W(50\%) & 9.0 & \multicolumn{1}{c|}{23.0} & 14.3 & \multicolumn{1}{c|}{25.6} & 10.7 & \multicolumn{1}{c|}{25.0} & 11.0 & \multicolumn{1}{c|}{25.5} & 11.3 & 24.8 \\ 
                           & \textbf{Context-I2W(100\%)} & \textbf{10.2} & \multicolumn{1}{c|}{\textbf{26.1}} & \textbf{17.5} & \multicolumn{1}{c|}{\textbf{28.7}} & \textbf{11.6} & \multicolumn{1}{c|}{\textbf{27.4}} & \textbf{12.1} & \multicolumn{1}{c|}{\textbf{28.2}} & \textbf{12.9} & \textbf{27.6} \\ 

\bottomrule
\end{tabular}}
\caption{Results on ImageNet for domain conversion.}
\label{tab:imgnet_add}
\end{table*}

% 0.85
\begin{table*}[t]
\centering
\scalebox{1.1}
{\scriptsize
\begin{tabular}{cccccccccccc}
\toprule
                            &                                                          & \multicolumn{2}{c}{Dress}                                                               & \multicolumn{2}{c}{Shrit}                                                      & \multicolumn{2}{c}{TopTee}                                                              & \multicolumn{2}{c}{Average}                               \\ \cmidrule(lr){3-4}\cmidrule(lr){5-6}\cmidrule(lr){7-8}\cmidrule(lr){9-10}
                            %\cline{3-10} 
Supervision                 & Methods                                                  & R10                                  & R50                                              & R10                         & R50                                              & R10                                  & R50                                              & R10                         & R50                         \\ \hline
                            & Image-only                                               & 5.4                                  & \multicolumn{1}{c|}{13.9}                        & 9.9                         & \multicolumn{1}{c|}{20.8}                        & 8.3                                  & \multicolumn{1}{c|}{17.7}                        & 7.9                         & 17.5                        \\
                            & Text-only                                                & 13.6                                 & \multicolumn{1}{c|}{29.7}                        & 18.9                        & \multicolumn{1}{c|}{31.8}                        & 19.3                                 & \multicolumn{1}{c|}{37.0}                        & 17.3                        & 32.9                        \\
                            & Image+Text                                               & 16.3                                 & \multicolumn{1}{c|}{33.6}                        & 21.0                        & \multicolumn{1}{c|}{34.5}                        & 22.2                                 & \multicolumn{1}{c|}{39.0}                        & 19.8                        & 35.7                        \\
                            & Pic2Word                                                & 20.0                                 & \multicolumn{1}{c|}{40.2}                        & 26.2                        & \multicolumn{1}{c|}{43.6}                        & 27.9                                 & \multicolumn{1}{c|}{47.4}                        & 24.7                        & 43.7                        \\
                            & SEARLE-XL                                               & 20.3                                 & \multicolumn{1}{c|}{43.2}                        & 27.4                                 & \multicolumn{1}{c|}{45.7}                        & 29.3                                 & \multicolumn{1}{c|}{50.2}                        & 25.7                        & 46.3                        \\
                            & Context-I2W(50\%)                                              & 21.4                                 & \multicolumn{1}{c|}{43.7}                        & 28.1                                 & \multicolumn{1}{c|}{46.9}                        & 29.7                                 & \multicolumn{1}{c|}{51.4}                        & 26.4                        & 47.3                        \\
                            & \textbf{Context-I2W(100\%)}                                             & \textbf{23.1}                        & \multicolumn{1}{c|}{\textbf{45.3}}               & \textbf{29.7}                        & \multicolumn{1}{c|}{\textbf{48.6}}               & \textbf{30.6}                        & \multicolumn{1}{c|}{\textbf{52.9}}               & \textbf{27.8}               & \textbf{48.9}               \\ 
                            
                            % \cmidrule(lr){1-10}
\multirow{-8}{*}{ZERO-SHOT} & \multicolumn{1}{c}{{\color[HTML]{9B9B9B} SEARLE-XL-OTI}} & {\color[HTML]{9B9B9B} 21.6} & \multicolumn{1}{c|}{{\color[HTML]{9B9B9B} 44.5}} & {\color[HTML]{9B9B9B} \textbf{30.4}} & \multicolumn{1}{c|}{{\color[HTML]{9B9B9B} 47.5}} & {\color[HTML]{9B9B9B} \textbf{30.9}} & \multicolumn{1}{c|}{{\color[HTML]{9B9B9B} 51.8}} & {\color[HTML]{9B9B9B} 27.6} & {\color[HTML]{9B9B9B} 47.9} \\
\bottomrule
\end{tabular}}
\caption{Results on Fashion-IQ for attribute manipulation.}
% \caption{Results on Fashion-IQ validation set. SEARLE-XL-OTI is the result reported by the authors without distillation.}
\label{tab:fashion_add}
\end{table*}

% 0.80
\begin{table}[t]
\centering
\scalebox{1.1}
{\scriptsize
\begin{tabular}{cccccccccccc}
\toprule
Supervision                                     & Methods                           & R1            & R5            & \multicolumn{1}{l}{R10} \\  \midrule
\multicolumn{1}{c|}{\multirow{6}{*}{ZERO-SHOT}} & \multicolumn{1}{c|}{Image-only}   & 8.6           & 15.4          & 18.9                    \\
\multicolumn{1}{c|}{}                           & \multicolumn{1}{c|}{Text-only}    & 6.1           & 15.7          & 23.5                    \\
\multicolumn{1}{c|}{}                           & \multicolumn{1}{c|}{Image+Text}   & 10.2          & 20.2          & 26.6                    \\
\multicolumn{1}{c|}{}                           & \multicolumn{1}{c|}{Pic2Word}     & 11.5          & 24.8          & 33.4                    \\
\multicolumn{1}{c|}{}                           & \multicolumn{1}{c|}{Context-I2W(50\%)}     & 12.1          & 25.6          & 34.4                    \\
\multicolumn{1}{c|}{}                           & \multicolumn{1}{c|}{\textbf{Context-I2W(100\%)}} & \textbf{13.5}          & \textbf{28.5}          & \textbf{38.1}           \\ 
\bottomrule
\end{tabular}}
\caption{Results of the object composition task using COCO.}
\label{tab:coco_add}
\end{table}

% 0.73
\begin{table}[t]
\centering
\scalebox{1.1}
{\scriptsize
\begin{tabular}{cccccccccccc}
\toprule
Supervision                                      & Methods                                                   & R1                                               & R5                                               & \multicolumn{1}{l}{R10}                          & R50                                              \\ \midrule
\multicolumn{1}{c|}{}                            & \multicolumn{1}{c|}{Image-only}                           & 7.4                                              & 23.6                                             & 34.0                                             & 57.4                                             \\
\multicolumn{1}{c|}{}                            & \multicolumn{1}{c|}{Text-only}                            & 20.9                                             & 44.8                                             & 56.7                                             & 79.1                                             \\
\multicolumn{1}{c|}{}                            & \multicolumn{1}{c|}{Image+Text}                           & 12.4                                             & 36.2                                             & 49.1                                             & 78.2                                             \\
\multicolumn{1}{c|}{}                            & \multicolumn{1}{c|}{Pic2Word}                             & 23.9                                            & 51.7                                           & 65.3                                            & 87.8                                            \\
\multicolumn{1}{c|}{}                            & \multicolumn{1}{c|}{SEARLE-XL}                   & 24.2                                            & 52.4                                            & 66.3                                            & 88.6                                   \\
\multicolumn{1}{c|}{}                            & \multicolumn{1}{c|}{Context-I2W(50\%)}                   & 24.8                                            & 53.6                                            & 67.1                                            & 88.9                                   \\
\multicolumn{1}{c|}{}                             & \multicolumn{1}{c|}{\textbf{Context-I2W(100\%)}}                         & \textbf{25.6}                                   & \textbf{55.1}                                   & \textbf{68.5}                                   & \textbf{89.8}                                   \\
\multicolumn{1}{c|}{\multirow{-8}{*}{ZERO-SHOT}} & \multicolumn{1}{c|}{{\color[HTML]{9B9B9B} SEARLE-XL-OTI}} & \multicolumn{1}{l}{{\color[HTML]{9B9B9B} 24.9}} & \multicolumn{1}{l}{{\color[HTML]{9B9B9B} 52.3}} & \multicolumn{1}{l}{{\color[HTML]{9B9B9B} 66.3}} & \multicolumn{1}{l}{{\color[HTML]{9B9B9B} 88.6}} \\

\bottomrule
\end{tabular}}
\caption{Results on CIRR for object composition.}
\label{tab:CIRR_add}
\end{table}

In Table \ref{tab:imgnet_add}-\ref{tab:CIRR_add}, we provide further evidence of the effectiveness and efficiency of Context-I2W. By training the model with only 50\% of the training data, we achieve results comparable to the state-of-the-art (SoTA), surpassing SoTA by an average margin of 0.73\% to 1.40\%. Furthermore, in comparing the CIRR and Fashion-IQ datasets, we contrasted our model with \textbf{SEARLE-XL-OTI}. This model, which performs better than SEARLE-XL without knowledge distillation, is considered less efficient due to its reliance on GPT during training. Specifically, when using a single A100 GPU, SEARLE-XL-OTI requires about 30 seconds to process a single image and around 1 second per image when using a batch size 256. However, Context-I2W achieves higher efficiency and better average performance than SEARLE-XL-OTI by 0.60\% to 1.73\%. This comprehensive analysis strongly validates both the effectiveness and efficiency of Context-I2W.

\subsection{D. More Visualization Experiments}

\begin{figure*}[t]
    \centering
    \centering
    \includegraphics[width=1.0\linewidth]{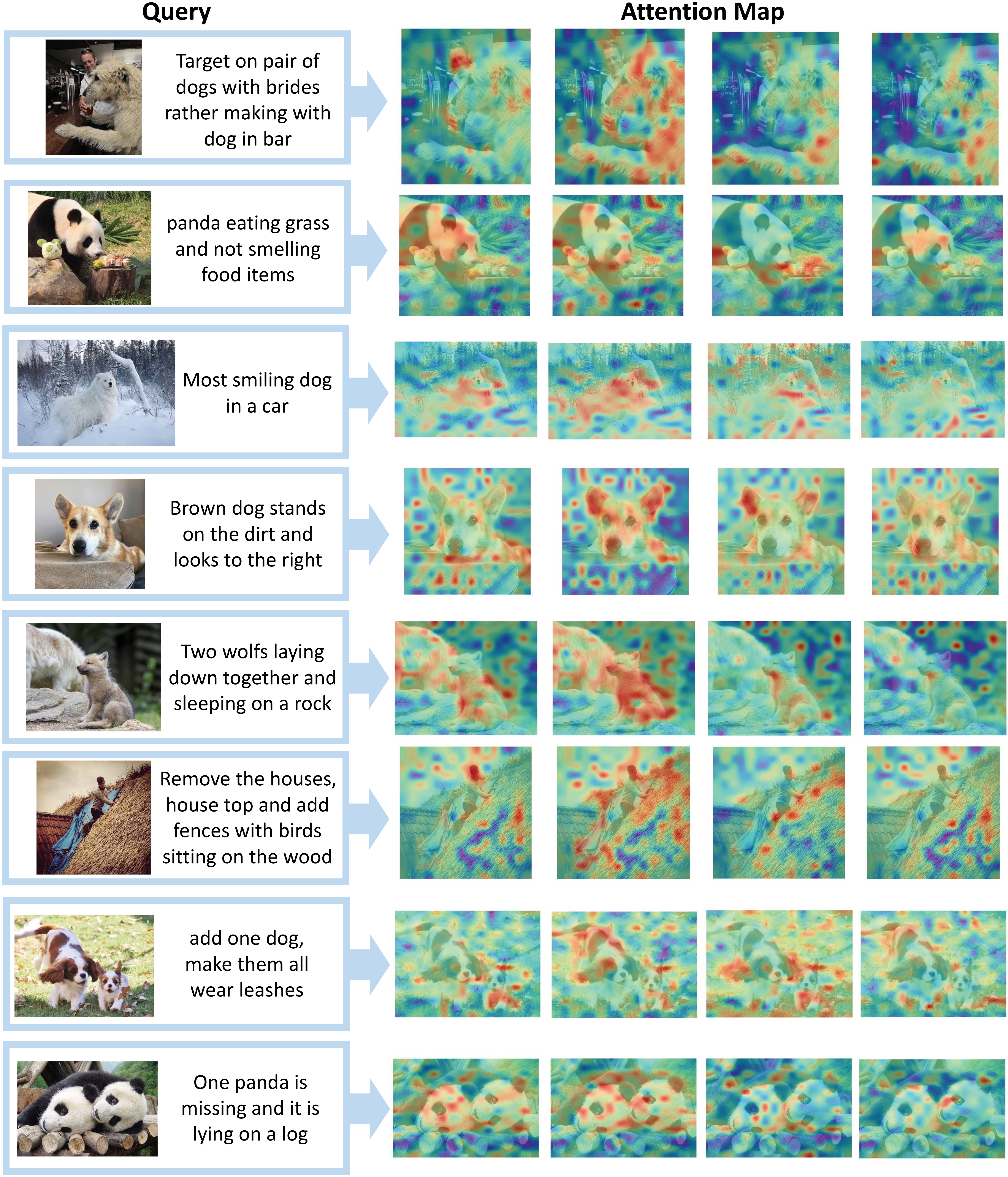}
    \caption{Attention map visualization of learnable queries.}
    % \vspace{-10pt}
    \label{fig:attn_add}
\end{figure*}

In Figure \ref{fig:attn_add}, we visualize more samples of the attention maps of each learnable query from the last block. The four queries obviously attend to different targets in the whole image: the first two queries mainly focus on the object attributes and foreground information. In comparison, the last two queries mostly consider the background and scene information. This pattern is usually consistent. These attention maps provide insight that our model is interpretable in extracting specific visual features of different targets, which supports fine-grained visual information selection for the context.

\subsection{E. More Implementation Details}
We adopt ViT-L/14 CLIP \cite{radford2021learning} pre-trained on 400M image-text paired data. For training Context-I2W, we utilize the Conceptual Caption dataset \cite{DBLP:conf/acl/SoricutDSG18}, which comprises 3M images. The number of cross-attention blocks is $6$, and each attention blocks with $8$ head. The dimension of hidden layers of 3-Layer MLP is $512$. The number of learnable queries is $4$. To improve training stability, we initialize the learnable scalar of tanh-gating to 0 \cite{bachlechner2021rezero}. We employ AdamW \cite{loshchilov2018decoupled} with a learning rate of $5\times10^{-5}$, weight decay of $0.1$, and a linear warmup of $10000$ steps. The batch size for contrastive learning is $1024$. Our model is trained on $4$ Tesla V100 (32G) GPUs for $24$ hours with $1$ hour for \textit{spacy} preprocessing. To ensure reliable results, we report the performance averaged over three trials. Moreover, we conduct ablation studies on CIRR test sets and FashionIQ validation sets. For FashionIQ, we consider the average recall. 

\subsubsection{E.1. More Dataset Details.} We further introduce our evaluation ZS-CIR datasets in four composed setups, object composition, domain conversion, object/scene manipulation, and attribute manipulation.

\noindent \textbf{(1) Domain conversion}. This setup evaluates the ability to compose real images and domain information to retrieve corresponding domain-specific images. We utilize ImageNet \cite{deng2009imagenet} and ImageNet-R \cite{Hendrycks_2021_ICCV}, which comprises 200 classes with diverse domains and has domain annotations. Following Pic2Word, we pick cartoon, origami, toy, and sculpture as the evaluation target to avoid noise in the annotations. With this selection, we have 16,983 images as candidates. In the evaluation, given the real image from ImageNet and target domain names, we compose the query following the procedure in (a) in the Inference section. \textit{e.g.,} \texttt{a cartoon of [REPLACE]}.

\noindent \textbf{(2) Object composition}.  We evaluation on the validation split (5000 images) of COCO \cite{10.1007/978-3-319-10602-1_48}, which dataset contains images with corresponding lists of object classes and instance mask of query images. Following Pic2Word, we randomly crop one object and mask its background using its instance mask to create a query for each image. The list of object classes is used as text specification. Given the reference image and class list, we compose a query by following (b) in the Inference section. \textit{e.g.,} \texttt{a photo of [REPLACE], [cat] and [dog]}.

\noindent \textbf{(3) Object/scene manipulation by text description}. In this setup, a reference image is provided alongside a text description containing instructions for manipulating either an object or the background scene depicted in the reference image. This composition of the reference image and text description enables the retrieval of manipulated images. We evaluation on the test split of CIRR \cite{Liu_2021_ICCV} using the standard evaluation protocol following previous works \cite{Saito_2023_CVPR, baldrati2023zero}, and query texts is composed following the procedure in (c) of the Inference section.

\noindent \textbf{(4) Attribute manipulation}. We employ Fashion-IQ \cite{Wu_2021_CVPR}, which includes various modification texts related to image attributes. These attribute manipulations are given as a sentence. As with CIRR, we adopt the standard evaluation protocol and create query texts following the procedure provided in (c) of the Inference section. In evaluation, we employ the validation set, following previous works \cite{Baldrati_2022_CVPR, Saito_2023_CVPR, baldrati2023zero}.

\end{document}